\documentclass[utf8]{article}
\usepackage{arxiv}
\usepackage{graphicx}
\usepackage[blocks]{authblk} 
\usepackage{natbib}
\usepackage{graphicx}
\usepackage{booktabs} 
\usepackage{tikz}
\usepackage{amssymb}
\usepackage{amsmath}
\usepackage{algorithm}
\usepackage[noend]{algpseudocode}
\usepackage{caption}
\usepackage{url,lineno,microtype,subcaption}
\usepackage{hyperref}
\hypersetup{
  colorlinks   = true, 
  urlcolor     = blue, 
  linkcolor    = blue, 
  citecolor    = blue  
}
\usepackage[onehalfspacing]{setspace}
\usepackage{soul}
\renewcommand\hl[1]{#1}

\definecolor{lime}{HTML}{A6CE39}
\DeclareRobustCommand{\orcidicon}{%
    \begin{tikzpicture}
    \draw[lime, fill=lime] (0,0)
    circle [radius=0.16]
    node[white] {{\fontfamily{qag}\selectfont \tiny ID}};
    \draw[white, fill=white] (-0.0725,0.1)
    circle [radius=0.007];
    \end{tikzpicture}
    \hspace{-2mm}
  }

\newcommand{\orcid}[1]{\href{https://orcid.org/#1}{\orcidicon}}

\title{Emergent communication enhances foraging behaviour in evolved swarms controlled by Spiking Neural Networks}
\author[1, $\dagger$]{\orcid{0000-0002-7239-434X}  Cristian Jimenez-Romero}
\author[1,4,$\dagger$,$\ast$]{\orcid{0000-0001-8869-215X } Alper Yegenoglu}
\author[1]{\orcid{0000-0001-6741-1435} Aaron Pérez Martin}
\author[1]{ \orcid{0000-0002-3168-5394} Sandra Diaz-Pier}
\author[1,2,3]{\orcid{0000-0001-6933-797X} Abigail Morrison}
\affil[1]{Simulation and Data Lab Neuroscience, Jülich Supercomputing Centre (JSC), Institute for Advanced Simulation, JARA, Forschungszentrum Jülich GmbH, Jülich, Germany}
\affil[2]{Institute of Neuroscience and Medicine (INM-6) and Institute for Advanced Simulation (IAS-6) and JARA BRAIN Institute I}
\affil[3]{Computer Science 3 - Software Engineering, RWTH Aachen University, Aachen, Germany}
\affil[4]{Institute of Geometry and Applied Mathematics, Department of Mathematics, RWTH Aachen University, Aachen, Germany}
\affil[$\dagger$]{shared authorship}
\affil[$\ast$]{contact: a.yegenoglu @ fz-juelich.de}
\date{}

\renewcommand{\headeright}{}
\renewcommand{\undertitle}{}
\renewcommand{\shorttitle}{Emergent Communication in Agents}

\hypersetup{
pdftitle={Emergent communication enhances foraging behaviour in evolved swarms controlled by Spiking Neural Networks},
pdfsubject={CS.NE},
pdfauthor={Alper Yegenoglu},
pdfkeywords={simulation, meta learning, high performance computing, connectivity generation, parameter exploration},
}

\begin{document}
\maketitle

\begin{abstract}
Social insects such as ants and termites communicate via pheromones which allows them to coordinate their activity and solve complex tasks as a swarm, e.g. foraging for food or finding their way back to the nest. 
This behaviour was shaped through evolutionary processes over millions of years. 
In computational models, self-coordination in swarms has been implemented using probabilistic or pre-defined simple action rules to shape the decision of each agent and the collective behaviour. 
However, manual tuned decision rules may limit the emergent behaviour of the swarm. 
In this work we investigate the emergence of self-coordination and communication in evolved swarms without defining any explicit rule.
For this purpose, we evolve a swarm of agents representing an ant colony. 
We use an evolutionary algorithm to optimize a spiking neural network (SNN) which serves as an artificial brain to control the behaviour of each agent.
The goal of the evolved colony is to find optimal ways to forage for food and return it to the nest in the shortest amount of time.
In the evolutionary phase, the ants are able to learn to collaborate by depositing pheromone near food piles and near the nest to guide other ants.
The pheromone usage is not manually encoded into the network; instead, this behaviour is established through the optimization procedure. 
We observe that pheromone-based communication enables the ants to perform better in comparison to colonies where communication via pheromone did not emerge. 
Furthermore, we assess the foraging performance of the ant colonies by comparing the SNN based model to a multi-agent rule based system. 
Our results show that the SNN based model can efficiently complete the foraging task in a short amount of time. 
Our approach illustrates that even in the absence of pre-defined rules, self coordination via pheromone emerges as a result of the network optimization.
This work serves as a proof of concept for the possibility of creating complex applications utilizing SNNs as underlying architectures for multi-agent interactions where communication and self-coordination is desired.

\end{abstract}

\section{Introduction}\label{sec:introduction}
Communication using chemical signals is an efficient way to interact and collaborate within swarm colonies.
For example, social insects such as ants and termites deposit pheromones to guide their cohorts while foraging for food. 
The coordination does not follow a top-down hierarchy, i.e., there is no directing leader. 
Instead, the coordination is self-organized and enables the swarm to solve complex tasks, such as finding the shortest path between the nest and the food source.
This behaviour is shaped through evolution over millions of years~\citep{wilson1967first, smith2008genetic,wilson2014natural, boudinot2022evidence}.

From a computational point of view the self-organized behaviour in swarms described in the literature has been implemented by using (simple) action rules which allow the agents to interact with other cohorts and the environment by depositing chemical signals.
The self-coordination of the swarm emerges as a result of these interactions, allowing the agents to solve problems, e.g.~finding the shortest path in an environment.
Defining explicit rules may help the system to solve specific tasks yielding in a foreseen behaviour, i.e.~the system is guided to find a certain solution.  
Although this approach may converge into a solution it may not be the most efficient one.
Moreover, defining rules, which lead to emergent interactions and self-coordination at the level of the swarm, may not be straightforward for complex tasks.
In this work we investigate the emergence of self-coordination and communication in evolved swarms without defining any explicit rules which describes the behaviour of each agent.
Specifically, our research questions are: 
\begin{itemize}
    \item Does  \textbf{self-organized coordination emerge} within the swarm, based on the simulated physiological attributes of the agents, the spike-based information processing of the controlling spiking neural network (SNN), and the environmental characteristics?
    \item Could the presence of a self-organized coordination between the agents \textbf{enhance the collaborative approach} to solve specific problems? 
\end{itemize}
To tackle these questions, we define a task and evolve an entire swarm of agents searching for food and bringing it back to the nest, resembling the ant colony foraging behaviour.
In particular, we are interested in exploring the role of stigmergy in the evolutionary emergence of communication within the swarm.
Stigmergy refers to indirect interactions which take place via the environment for the purpose of navigation, communication or collaboration.

\subsection{Related work}\label{subsec:related_work}
Ant colony foraging is a self-organized behaviour which is well investigated within the literature.
In these works several computational models replicate to some extent the dynamics observed in biological colonies.

For instance, the authors of~\citet{vittori2004modeling} and similarly of~\citet{bandeira2008modeling} compare the foraging behaviour of the Argentine ants Linepithema with an implementation of a probabilistic model.
Their models are able to find the shortest paths in a maze to a food source relying on self-organization based on pheromone trails. In \citet{hecker2015beyond}, the authors show the result of implementing an ant swarm algorithm to control simulated and real robots in environments, where the distribution of food changes.
They control each robot using a rule-based probabilistic model mimicking the foraging behavior of seed-harvester ants and optimize the performance of the collective behaviour using a genetic algorithm.
They analyze how the collective strategies and communication are evolved. The model presented in~\cite{wilensky1997netlogo} illustrates self-coordination and collaboration between ants through the use of pheromone.
This model is utilizing simple, pre-defined rules to control the local decisions of the ants. 

In an alternative approach, ~\citet{duan2014swarm} propose a framework to soft control agents, called shills, which are able to follow designed update rules.
Their task is based on an iterated prisoner's dilemma game, where shills can update their strategies using particle swarm optimization (PSO) mechanisms.
In their setting, shills can choose better performing strategies and provide those to other individuals.
The frequency of the communication between the shills can be steered by different parameter configuration.

Swarms have also been combined with spiking neural networks (SNNs) in the past. 
In~\citet{chevallier2010spikeants}, the authors implement a spatio-temporal model called "SpikeAnts" and depict the emergence of synchronized activities in an ant colony.
They describe a sparsely-connected network of spiking neurons, where each ant is represented by two interconnected neurons.
The network does not receive any external stimuli and does not incorporate learning rules. 
The emergent behaviour shows a distributed learning process in the population of agents and it is concluded that in their setting the swarms is able to self-organize. 

\hl{Artificial neural networks demonstrated to be a successful technique for controlling swarms. 
To optimize the architecture of these networks, evolutionary algorithms are often employed.}
For example,~\citet{christensen2006evolving} \hl{evolve a swarm of robots that performs hole-avoidance and phototaxis. In their work they compare different evolutionary algorithms, and conclude that evolutionary strategy performs best. Subsequently, they implement this technique on a group of three s-bot robots.}

Similarly,~\citet{trianni2009self} \hl{use evolutionary algorithms to study the design of self-organizing behaviour in swarm robotics. In their simulations, they run 20 independent instances of the design process. Following a similar approach to Christensen and Dorigo they apply the best instance on three s-bot robots.}

\citet{waibel2009genetic} \hl{examine how various types of selection pressure (individual vs. collective) and swarm composition (homogenous vs. heterogeneous) affect the performance in a foraging task. Their investigation involves three task variants and 4 combinations of selection pressures/team compositions, tested through 20 independent runs using both simulation and real robots (Alice robots).} 

In a more recent study,~\citet{ericksen2017automatically} \hl{utilize the Neuroevolution of Augmented Topologies (NEAT) algorithm to automate the design of a neural network controller which controls a swarm of homogeneous robots. They compare their system, called NeatFA (NEAT Foraging Algorithm), to existing swarm foraging algorithms, such as the Central Place Foraging Algorithm (CPFA), and the Distributed Deterministic Spiral Algorithm (DDSA). In their experiments involving large swarms, their evolved network outperforms the CPFA and DDSA when applied to a foraging task.}

\subsection{Our contribution}\label{subsec:contribution}
\hl{A significant number of the aforementioned methods for controlling swarms rely on engineered probabilistic models or rule-based systems to steer swarm interactions.
While these methods successfully induce collaborative behavior within the swarm, they pose a challenge as designers often drive this process through iterative experimentation, relying on their intuition and prior experiences}~\citep{francesca2016automatic}.
\hl{Trying to formalize all the rules and interactions which address not only individual but also collective behaviours in a dynamic realistic environment becomes complex. This approach may also not scale well if the complexity of the task or the number of agents increases. This predicament is known in the literature as the design problem}~\citep{trianni2008evolution}.
\hl{In contrast, recent efforts have aimed to address the design challenge by adopting connectionist models, wherein the swarm controller is built upon artificial neural networks.
Furthermore, another promising solution may lie in the application of spiking neural networks.
SNNs offer several distinct advantages, including sparse coding, precise spike timing encoding, and local plasticity rules. 
Sparse coding allows for efficient and economical representation of information within the network, reducing computational demands}~\citep{yamazaki2022spiking}. 
\hl{Local plasticity rules facilitate bio-inspired adaptive learning mechanisms within the network, potentially allowing swarms to adjust and optimize their responses to changing environmental conditions. 
SNNs have the capability to control agents and robots in dynamical environments}\citep{nichols2010case,beyeler2015gpu,nichols2012biologically}. 
\hl{Moreover, SNNs hold the potential to leverage specialized neuromorphic hardware, which is designed to emulate the brain's neural architecture efficiently. 
This hardware can significantly enhance the computational efficiency and real-time processing capabilities of swarm controllers based on spiking neural networks, further improving their overall performance in complex and dynamic environments}~\citep{basu2022spiking,ottati2023spike,putra2023topspark}.

\hl{In our approach, the agents are controlled by SNNs and are not strictly bound by predefined rules to guide their behavior. Instead, we allow  the optimization process the autonomy to explore solutions based on an objective fitness function to maximize the fitness measure.}
Each agent is controlled by an SNN composed of several neurons.
We optimize the synaptic weights and spike time delays of the entire network using genetic algorithms, so that the network is able to control the agents to navigate in the virtual environment while searching and collecting food and bringing it back to the nest.
At the beginning of the evolutionary process, the SNN is agnostic of the foraging task and there is no pre-defined communication or coordination mechanism encoded into the architecture of the SNN. 
There is no explicit mapping between the sensory input and the actions taken by the ants, e.g. visual information and pheromone depositing.  
\begin{figure}[t]
  \centering
  \includegraphics[width=0.46\columnwidth]{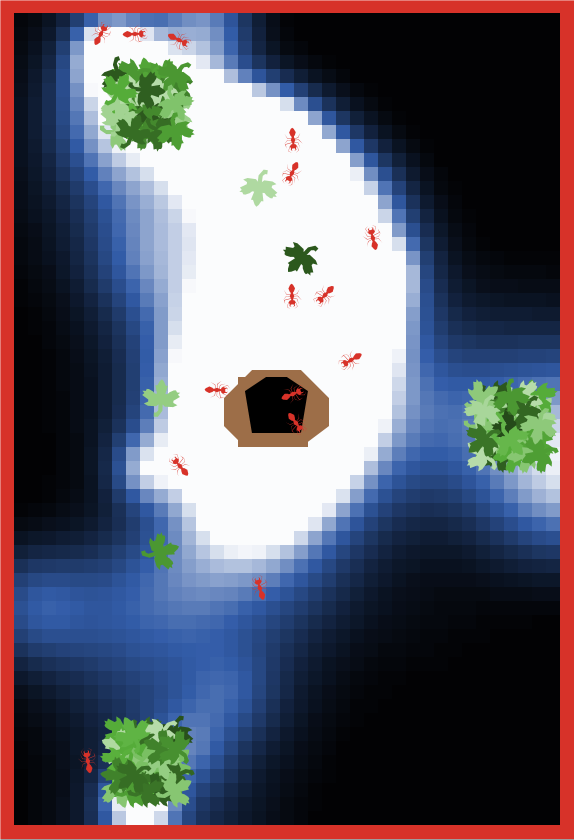}
  \caption{\label{fig:pheromone_concentration_teaser} Screenshot of the environment. The picture shows an ongoing simulation with an active foraging swarm. White/blue patches indicate the presence of pheromone. The brighter the color the higher the pheromone concentration. Food piles are depicted with green patches and leafs. The black hexagon in the middle is the nest. \hl{An impassable red wall is set around the environment.}}
\end{figure}
The virtual environment, in which the simulations are run, is based on the ant-colony model by~\citet{wilensky1997netlogo}.
In this model, the diffusion and evaporation of pheromone is simulated in a 2-dimensional grid world. 
For an example of a possible distribution of pheromone for an active swarm, see \autoref{fig:pheromone_concentration_teaser}. The ants colored in red are exploring the world for food (green leaves) and depositing pheromones near the food patches and in the vicinity of their nest.
While some ants are following the pheromone trail leading to the food patches or to the nest, others are exploring the environment.
Every ant is steered by a spiking neural network.
To optimize the network parameters and create the environment, we connect \hl{our own implemented} Learning to Learn framework~\citep[L2L;][]{yegenoglu2022exploring} with the multi-agent simulator NetLogo~\citep{tisue2004netlogo} and the spiking neural network simulator NEST~\citep{gewaltig2007nest}. 

\subsection{Structure of our manuscript}\label{subsec:structure}
Our manuscript is structured as follows.
First, in~\autoref{sec:methods}, we introduce our toolchain and setup of the environment.
We describe the experimental design, the structure of the agent network and how we employ a genetic algorithm to optimize the synaptic weights and delays over multiple generations.
Then, in~\autoref{sec:results} we analyze the emergent communication via pheromones and compare our network structure to the base model implemented in~\citet{wilensky1997netlogo} in terms of foraging performance.
We examine how the pheromone usage and concentration is shaped by the evolutionary process which enhances the self-coordination and communication between the ants.  
Moreover, we investigate the correlation between the input and output spike activity produced by an agent's network. 
This analysis shows us that certain input actions lead to specific output reactions.
We measure the significance and relate these actions to each other. 
Finally, we discuss the results of the swarm simulations, with special interest on the emergent communication, and we mention some avenues for future work. 

As a summary, in this manuscript we show that agents \hl{controlled by SNNs and} without pre-defined rules can self-organize through evolutionary optimization which allows them to solve complex tasks as a swarm, leveraging their physiological capabilities.
Interestingly, in our specific use case of an ant colony, we observe that the pheromone acts as an attractor rather than an event-based signalling mechanism, which resembles the behaviour in biological colonies.
This work provides a proof of concept environment which can be used to study learning and emergence of complex behaviour in multi-agent settings.

\section{Material and Methods}\label{sec:methods}
\begin{figure}[ht]
    \centering
    \includegraphics[width=1.\columnwidth]{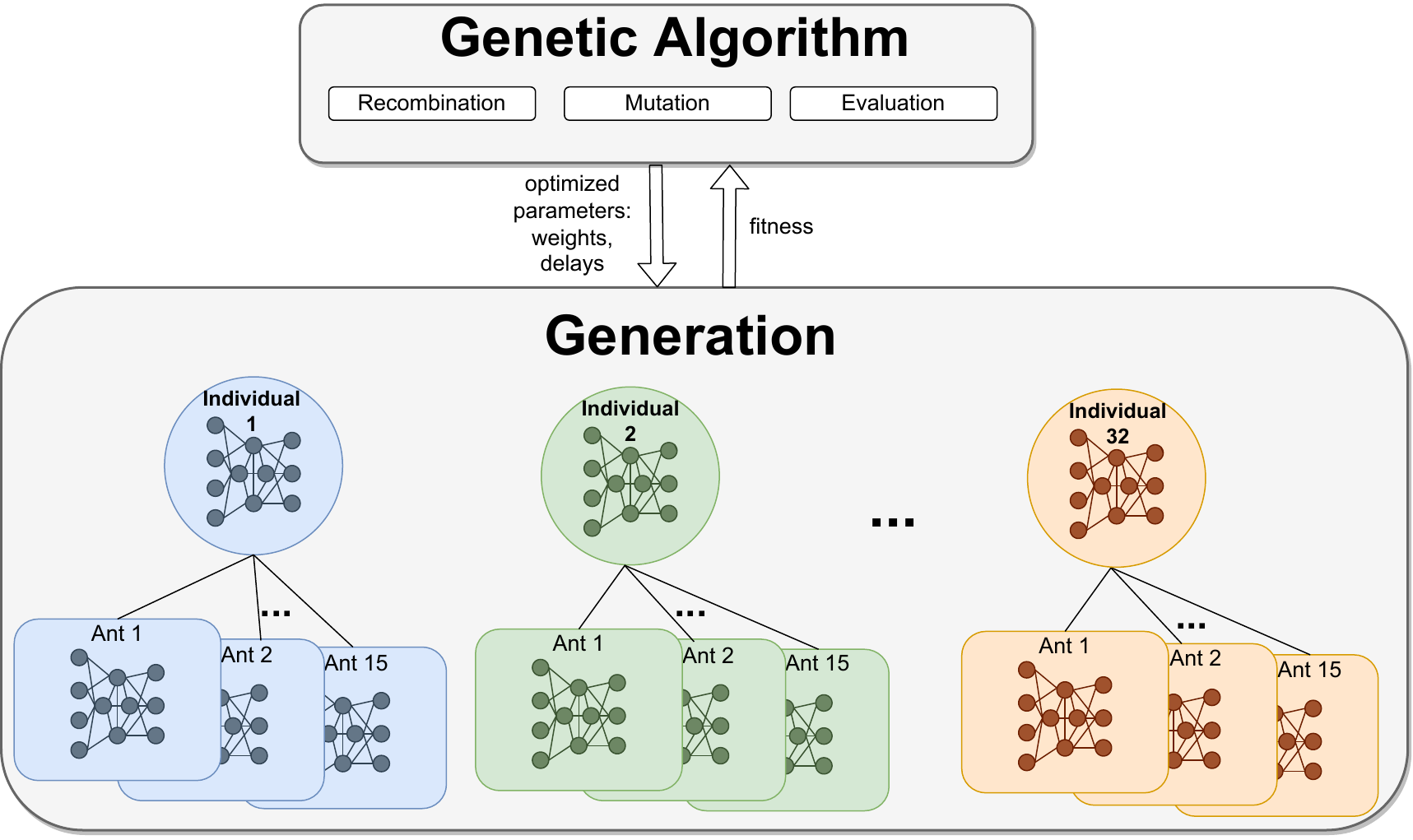}
    \caption{Optimization workflow. In each generation $32$ individuals are simulated. Each individual corresponds to an instance of an SNN and consists of $15$ ants with a copy of the same network. The genetic algorithm requires the calculated fitness of every individual to perform the optimization on the connection weights and delays.}
    \label{fig:sim_flow}
\end{figure}
We carry out optimization of a spiking neural network to steer an ant colony foraging for food using a genetic algorithm (GA).
The optimization workflow is depicted in~\autoref{fig:sim_flow}. 
There are $32$ individuals per generation.
Each individual corresponds to an instance of an SNN (see \autoref{fig:colony_net})
It is important to note, that an individual refers to the SNN instance and not to a simulated ant. 
There are $15$ ants per simulation forming the colony and every ant is controlled by an SNN, which is a copy of the same individual (network).

Since there are $32$ individuals we are running $32$ simulations in parallel.
Thus, we execute $15$ identical networks during one simulation per individual and $480$ networks in total per generation.
Each colony is simulated within a 2D environment to perform the foraging task.
After the simulation finishes, a fitness value is calculated to evaluate the performance of the run.

\hl{The GA orchestrates an exploratory process within the parameter space, specifically focusing on the weight and delay parameters associated with one-to-one synaptic connections between neurons in the network. Both parameters are instrumental in influencing the information transmission and processing dynamics within the neural network. 
The entirety of the weight-delay pairs associated with each connection in the network constitutes the genome of each individual within the population. Over successive generations and guided by mechanisms of selective pressure, the genetic algorithm iteratively fine-tunes the synaptic parameter values. This iterative process leads to the convergence of a configuration that maximizes the efficacy of the SNN with respect to the fitness function. An overview of the settings related to the optimization of the parameters can be found in the}~\autoref{sec:appendix}.

\subsection{Toolchain}\label{subsec:toolchain}
Our main toolchain\footnote{Our simulation and analysis scripts are publicly available at \url{https://github.com/Meta-optimization/emergent_communication_in_agents}} consists of three applications: The meta-learning framework Learning to Learn~\citep[L2L;][]{yegenoglu2022exploring}, the multi-agent simulator NetLogo~\citep{tisue2004netlogo} and the spiking neural network simulator NEST~\citep{gewaltig2007nest}.

\textbf{L2L} (Version 1.0.0-beta) is the orchestrator of the workflow (see~\autoref{sec:appendix} for L2L optimization details).
The framework requires the definition of the optimization algorithm, including its (hyper-)parameters, and the simulation whose parameters are going to be optimized.
The software architecture consists of a two loop structure, the inner loop (optimizee) and the outer loop (optimizer).
In the inner loop, the optimizee -- for example a neural network with or without learning capabilities -- is trained on a family of tasks.
In our setting, the SNNs in the inner loop do not learn and only the performance on the foraging task is evaluated, the optimization is conducted in the outer loop. 
The performance of the training or simulation run is evaluated by a fitness function, yielding a fitness which can be a scalar or a vector value.
In the outer loop, the (hyper-) parameters of the optimizee are optimized.
For example, if the optimizee is a network, the parameters could be the learning rate and the weights.
The aim is to reach an overall good performance and to enable faster learning on new, unknown tasks.
In this work we use the L2L framework's genetic algorithm as optimizer in the outer loop, which wraps the implementation from the DEAP~\citep{DEAP_JMLR2012} library.

\textbf{NetLogo} (Version 6.2.0) is a multi-agent simulator. 
Every object in a NetLogo simulation is an agent and can be used as means of communication with other agents.
With NetLogo it is easy to replicate the same simulation and agent.
In our settings the simulations are spiking neural networks and the agents are ants. 
Each ant is controlled by one instantiation of a common spiking neural network.
To communicate with each other the ants can deposit pheromones.
Patches are used to represent the pheromone which is dropped by the ants on the environment.
Each ant is also capable of sensing the presence of pheromones. 
NetLogo allows the simulation of the pheromone dynamics through its embedded language primitives.

The simulations uses the spiking neural simulator \textbf{NEST} (Version 3.0) as the back-end. 
NEST is a spiking neuronal simulator for biologically plausible, large-scale neural network simulations.
Simple neuron models with biological realistic connection structures can be simulated in an efficient way in a variety of hardware from local machines up to High Performance Computing (HPC) systems~\citep{jordan2018extremely}. 
NEST supports multi-threading and parallel execution with the message passage interface (MPI).

\subsection{Ant colony foraging for food}\label{subsec:ant_colony}
The task of the ant colony is to search for food and return it to the nest. 
In order to achieve this, each ant is controlled by its own network, a copy of the individual SNN.
In the following sections we describe the experimental setup  of the NetLogo environment (\autoref{subsubsec:experimental_env}), the neural model as well as the network architecture (\autoref{subsubsec:network_architecture}) and the sensory and actuator mechanisms (\autoref{subsubsec:sense_mechanism}) for the ants. 
Finally, we describe the colony fitness function (\autoref{subsubsec:fitness_calc}). 

\subsubsection{Experimental NetLogo environment}\label{subsubsec:experimental_env}
\hl{The experimental environment employs a grid-based system, where the world is represented as a discrete grid comprising individual cells. Within this setup, the agents are limited to navigating from one grid cell to another. The agent's behavior during each simulation step is determined by the output of the controlling network, allowing it to move forward by one cell at the time and providing the option to rotate either clockwise or anticlockwise.
Simultaneously, the pheromone dynamics also operate on the same grid as the agents' navigation. The pheromone levels are updated and influenced by the agents' actions and interactions with the environment. As the agents move through the grid, they deposit and sense pheromone levels at their current positions, influencing their decision-making processes and potentially affecting the behavior of neighboring agents.
Furthermore, the pheromone in this environment is subjected to two additional processes: diffusion and evaporation. Diffusion allows the pheromone to spread from the cells where it was initially deposited to neighboring cells, leading to a gradual spread and potential concentration gradients. On the other hand, evaporation causes the pheromone levels in each cell to decrease over time, ensuring that old pheromone trails gradually diminish and do not persist indefinitely.
The combination of agent movement, pheromone deposition, diffusion, and evaporation in this grid-based environment creates a dynamic and evolving system. These processes play a crucial role in shaping the overall behaviour and emergent patterns observed in the model.}

\subsubsection{Neural model and network architecture}\label{subsubsec:network_architecture}
\begin{figure}[t]
    \centering
    \includegraphics[width=0.5\columnwidth]{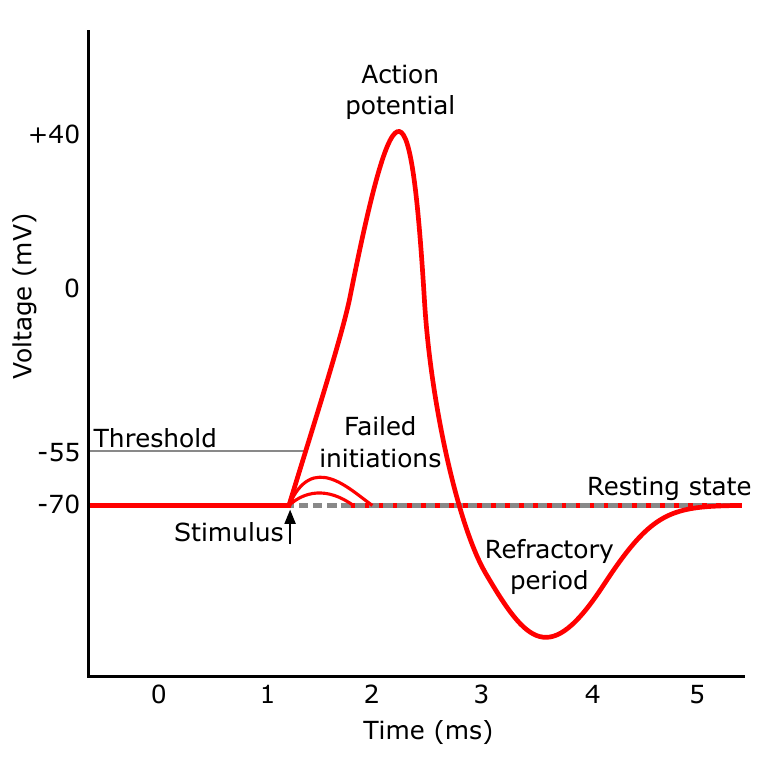}
    \caption{Action potential of a neuron. Incoming currents progressively raise the membrane potential of a neuron until the firing threshold is reached, then a spike is initiated, marking a firing event. Subsequently, the neuron enters an absolute refractory state, during which the membrane potential stabilizes at the resting potential. Throughout this phase, the neuron remains unresponsive to incoming stimuli until the refractory period concludes. Adapted from Wikimedia Commons.}
    \label{fig:action_potential}
\end{figure}

\hl{NEST is employed to build the SNN and to simulate neural dynamics. Additionally, it simulates the application of electrical stimulation to the neurons through the utilization of Direct Current (DC) generators. Each neuron within these networks is represented by a leaky integrate-and-fire (LIF) model featuring alpha function-shaped synaptic currents} \citep{gerstner2014neuronal}. 
\hl{Neurons communicate via action potential or spikes, a visualization of this sequence is depicted in}~\autoref{fig:action_potential}. \hl{The pertinent LIF parameters of our model can be found in}~\autoref{tab:lif_parameters} in~\autoref{sec:appendix}.

\begin{figure}[ht]
    \centering
    \includegraphics[width=0.9\columnwidth]{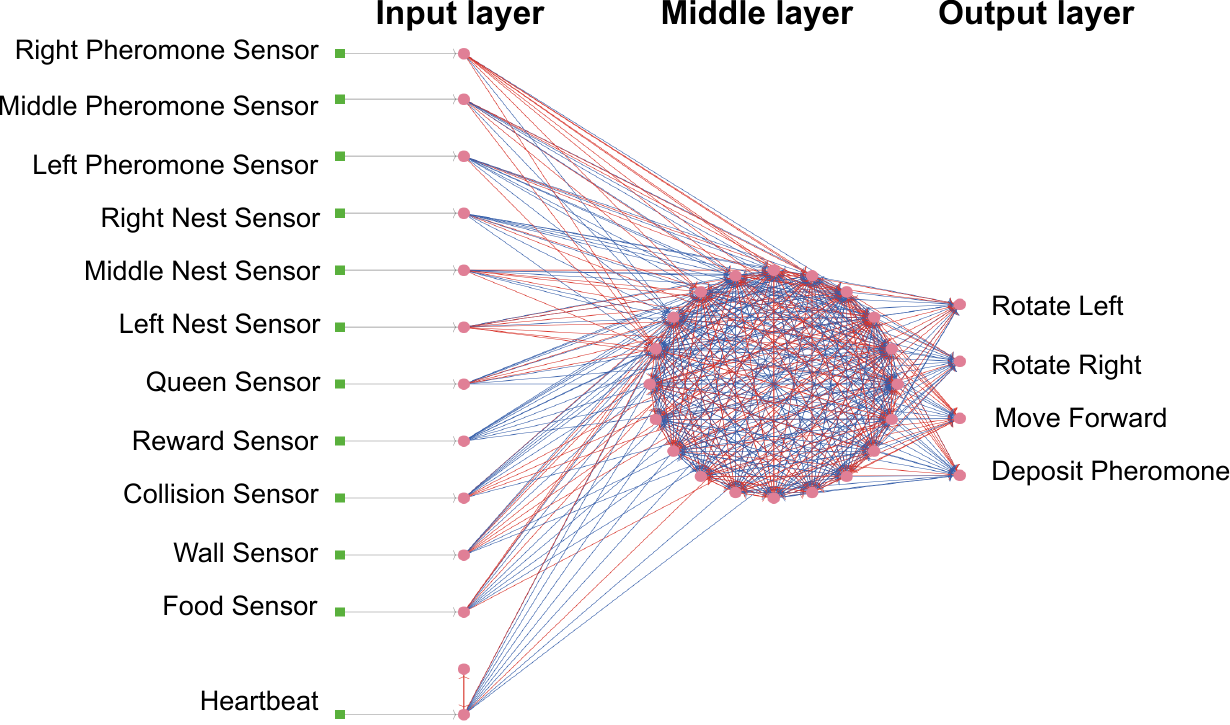}
    \caption{Spiking network to control an ant of the colony. The topology comprises $12$ neurons for the input, $20$ all-to-all connected neurons in the middle and four neurons in the output.}
    \label{fig:colony_net}
\end{figure}

\hl{The overall structure of the SNN can be segmented into three distinct layers. Firstly, the input layer assimilates sensory data that an ant can perceive, for example, encompassing sensors designed for detecting pheromone signals, the scent of the nest, reward or nociceptive stimuli. Secondly, the middle layer is interconnected in an all-to-all manner. Lastly, the output layer reflects the available actions an ant can execute, including forward movement, rotation and depositing pheromone.

Each layer maintains connections with the subsequent layer through a comprehensive all-to-all connectivity. This architecture allows the optimization algorithm to discern the most efficient configuration for the SNN, tailored to the task at hand, without imposing any constraints.}

In our setup each ant is controlled by its own network, a copy of the individual SNN.
An example of such a network is shown in \autoref{fig:colony_net}. 
The network consists of $36$ neurons and $720$ connections.

\subsubsection{Sensory and actuator mechanisms}\label{subsubsec:sense_mechanism}
\begin{figure}[ht]
    \centering
    \includegraphics[width=0.75\columnwidth]{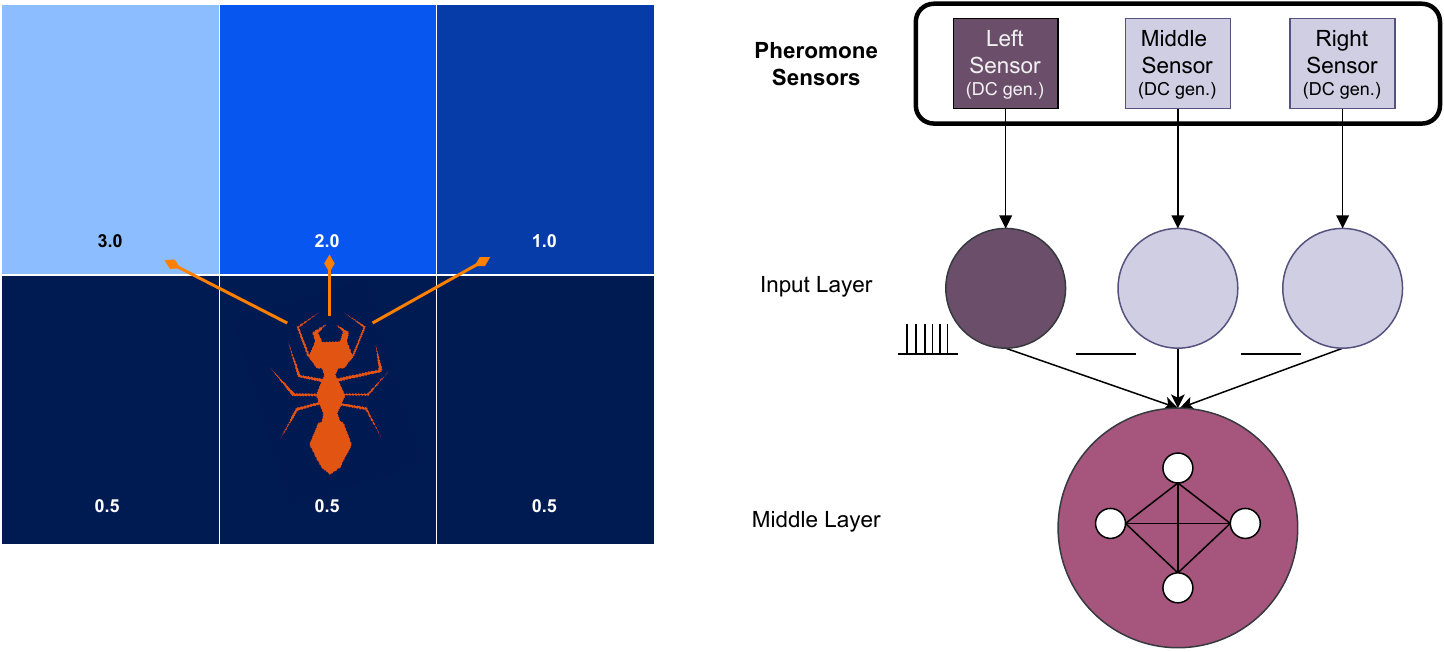}
    \caption{Pheromone detection mechanism of the ant. The left side illustrates how the ant assesses the pheromone concentration in the three neighboring cells directly ahead. On the right side, each sensor corresponds to a distinct neuron. The sensor that detects the highest pheromone concentration will activate its corresponding neuron, while the other neurons remain inactive. If the pheromone concentration is identical in at least two of the cells, the neuron corresponding to the middle sensor is activated. This scenario accounts for cases where the ant encounters a relatively uniform distribution of the pheromone scent across the adjacent cells. }
    \label{fig:pheromone_sen}
\end{figure}
\begin{figure}[ht]
    \centering
    \includegraphics[width=0.75\columnwidth]{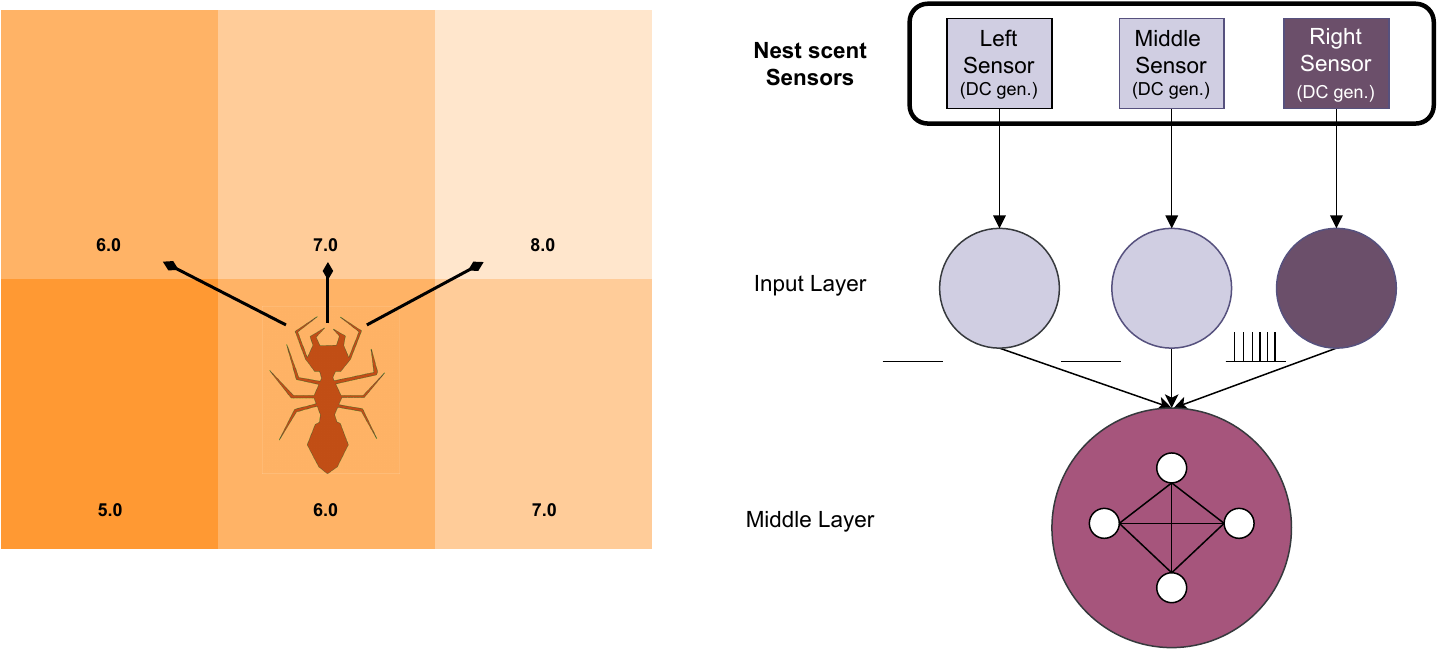}
    \caption{Nest detection mechanism of the ant. Similar to the ant's mechanism for detecting pheromones, the process of nest detection involves the agent's sensory system assessing the concentration of a distinct scent associated with the location of a nest. This scent, is emitted by the nest area itself.
    The ant employs sensors dedicated to perceiving this nest scent in the three adjacent cells directly ahead of its current position as shown on the left side. Each of these sensors corresponds to a unique neuron within the agent's neural network as illustrated on the right side. 
    Similar to the pheromone detection mechanism, the sensor that detects the highest concentration of the nest scent will activate its corresponding neuron. Meanwhile, the other neurons remain inactive, preserving the specificity of the sensory response.
    In situations where the concentration of the nest scent is identical in at least two of the cells, the neuron linked to the middle sensor is triggered. }
    \label{fig:nest_sen}
\end{figure}

\hl{To encode the diverse sensory inputs, encompassing various stimuli such as food, walls, pheromones, nest scent, reward, and collision, originating from different sensor types, a direct current (DC) generator is linked to each input neuron. When the associated sensor is activated, the DC generator emits a consistent DC signal. Consequently, the input neurons that are activated can sustain a high-rate spike train as long as the corresponding sensory stimulus persists.}
These sensors are connected to the twelve neurons in the input layer.
The top three neurons  in \autoref{fig:colony_net} are sensory neurons which react to the direction of the pheromone. 

\begin{figure}[h]
    \centering
    \includegraphics[width=0.75\columnwidth]{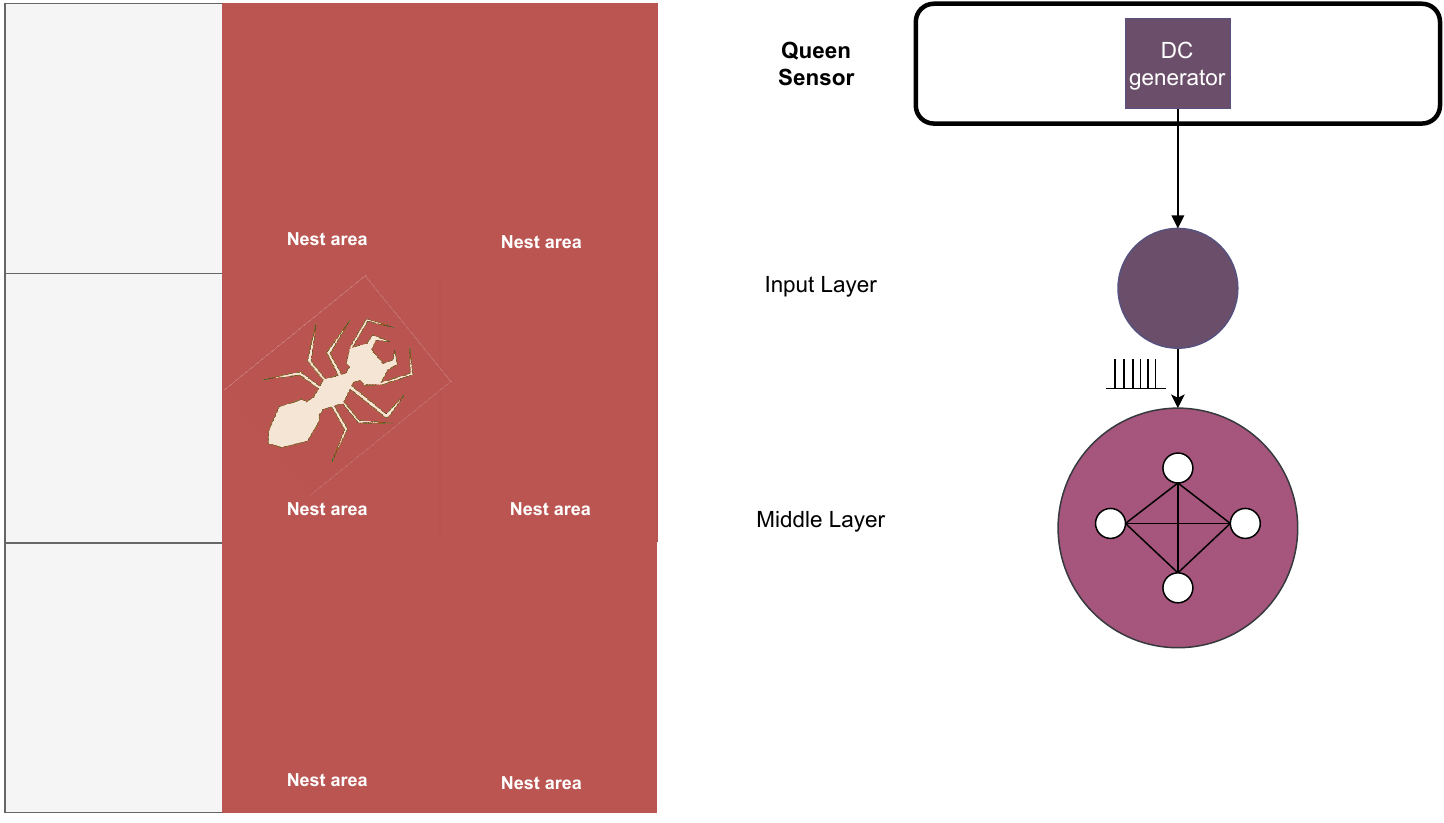}
    \caption{The Queen sensor is activated whenever the agent is situated within the nest area as shown on the left side. Upon activation, as depicted on the right side, it initiates the firing activity of its corresponding input neuron, maintaining this activity as long as the agent remains within the nest area.}
    \label{fig:queen_sen}
\end{figure}
\begin{figure}[ht]
    \centering
    \includegraphics[width=0.75\columnwidth]{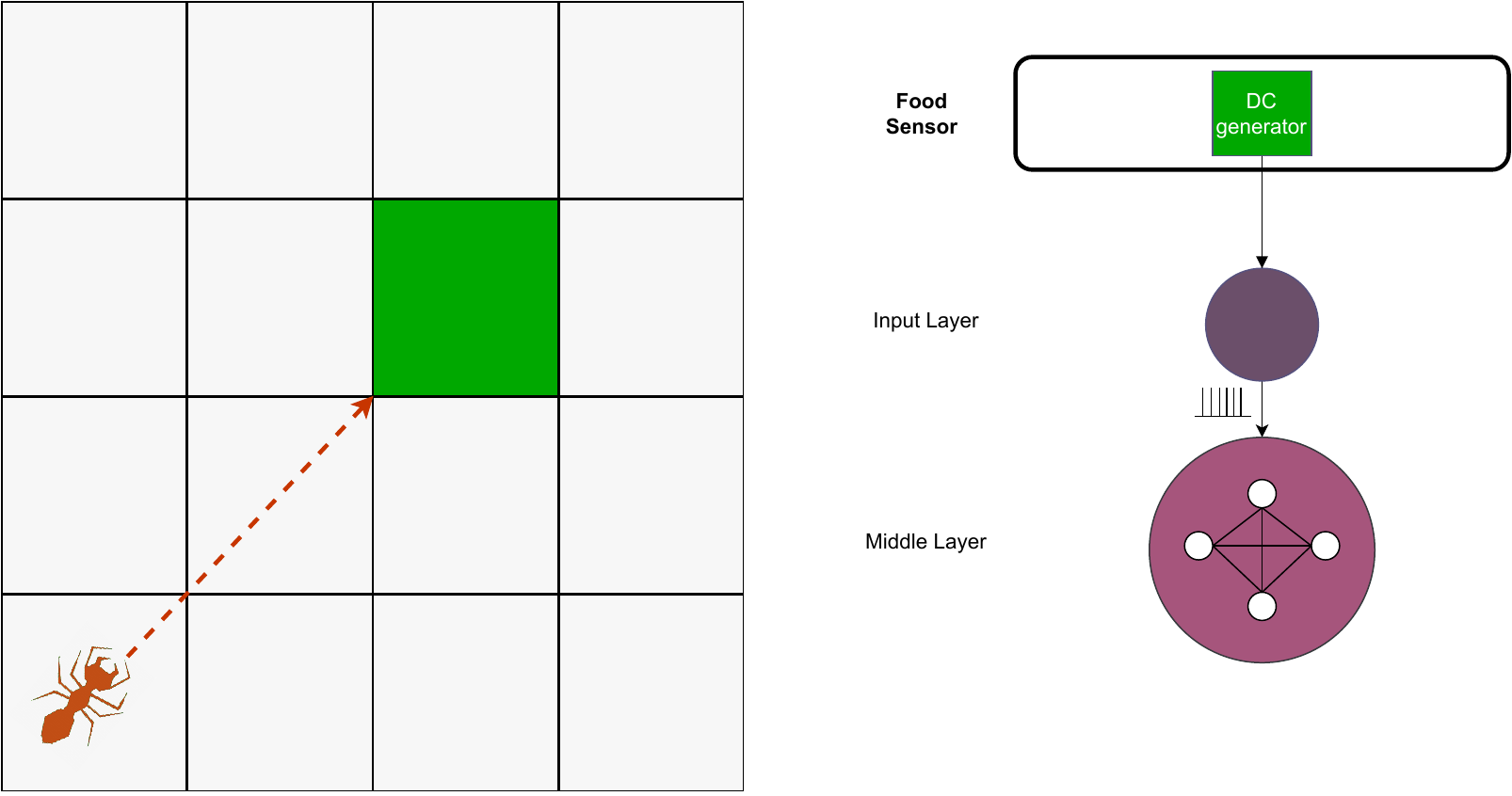}
    \caption{Food detection mechanism. Food sources are depicted as green-colored patches in the environment. As shown on the left side the sensor detects food located 2 patches directly in front of the ant. This detection initiates the activation of the corresponding neuron, which is illustrated on the right side.}
    \label{fig:food_sen}
\end{figure}
\begin{figure}[ht]
    \centering
    \includegraphics[width=0.75\columnwidth]{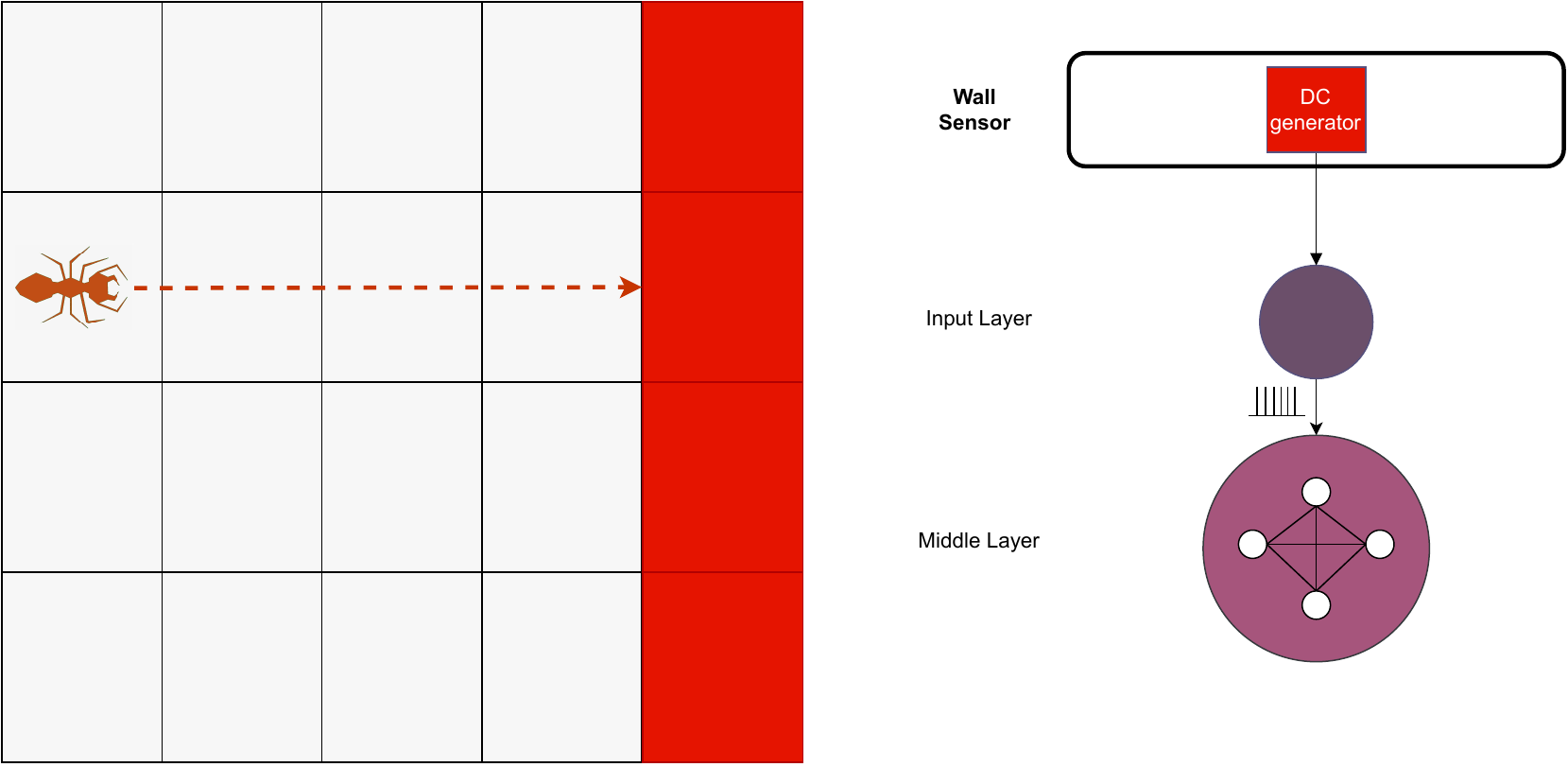}
    \caption{Wall detection mechanism. Walls are represented by red-colored patches. The sensor detects a wall located 4 patches directly in front of the ant as depicted on the left side. This detection initiates the activation of the corresponding neuron as illustrated on the right side.}
    \label{fig:wall_sen}
\end{figure}

The pheromone detection mechanism is elucidated in~\autoref{fig:pheromone_sen}.
The next three neurons can locate the nest scent as depicted in \autoref{fig:nest_sen}.
The queen sensor is activated when the ant is within the \hl{area} of the nest, see~\autoref{fig:queen_sen}.
The reward and punishment for the ant are determined by the reward and collision sensors.
\hl{The reward sensor is activated when the ant touches the food or returns it to the nest and the collision sensor is active whenever the ant collides with the red wall.}
\hl{The food and wall sensors enable the ant to detect 
objects within a range extending up to 5 patches directly ahead of it, this is exemplary illustrated in}~\autoref{fig:food_sen} and~\autoref{fig:wall_sen}. 
\hl{Each of these sensors is connected to a neuron that, when the sensor is activated, initiates firing activity within the associated neuron.}

The heartbeat neuron stimulates the network in every timestep by supplying a small direct current in order to maintain activity in the network.

Each ant can only collect one food patch at a time and must transport it back to the nest before looking for food again.
There is no pre-programmed behaviour to signal the ants that this is the activity pattern they should follow. 
This pattern is enforced by not allowing ants to carry more than one piece of food and rewarding ants when they bring the food to the nest, at which point they also regain the capability to pick up another piece of food.
In every generation we change the position of the nest randomly, in order to avoid overfitting the network and make the colony more adaptable to the location of the food piles. 

As depicted in \autoref{fig:pheromone_concentration_teaser}, every agent can deposit  pheromone which is represented by a white patch.
The dropped pheromone spreads from the point (patch) of release to a certain amount of neighboring patches according to the diffusion rate, defined in the simulation parameters.
Pheromone concentration also decreases (evaporates) exponentially over time.
There is no predefined behaviour that binds depositing pheromone to any action or sensor. 
Other agents can smell the pheromone and react to it. 
The response elicited in an agent by the pheromone emerges from the evolutionary process carried out by the genetic algorithm. 
For example, the sensing of pheromone at a certain location could modify the trajectory of an agent looking for a food source or the nest.

\hl{When an agent is positioned directly over a food  patch, the reward sensor triggers spikes in the corresponding reward  neuron. Concurrently, the amount of food within that patch diminishes by  a single unit. Furthermore, this activity temporarily restricts the  agent's ability to decrease the food count in other patches. The  restriction remains in place until the agent returns to the nest area,  restoring the agent's capability to engage with and decrease the food  count in other food patches.}


\begin{figure}[t]
    \centering
    \includegraphics[width=0.8\columnwidth]{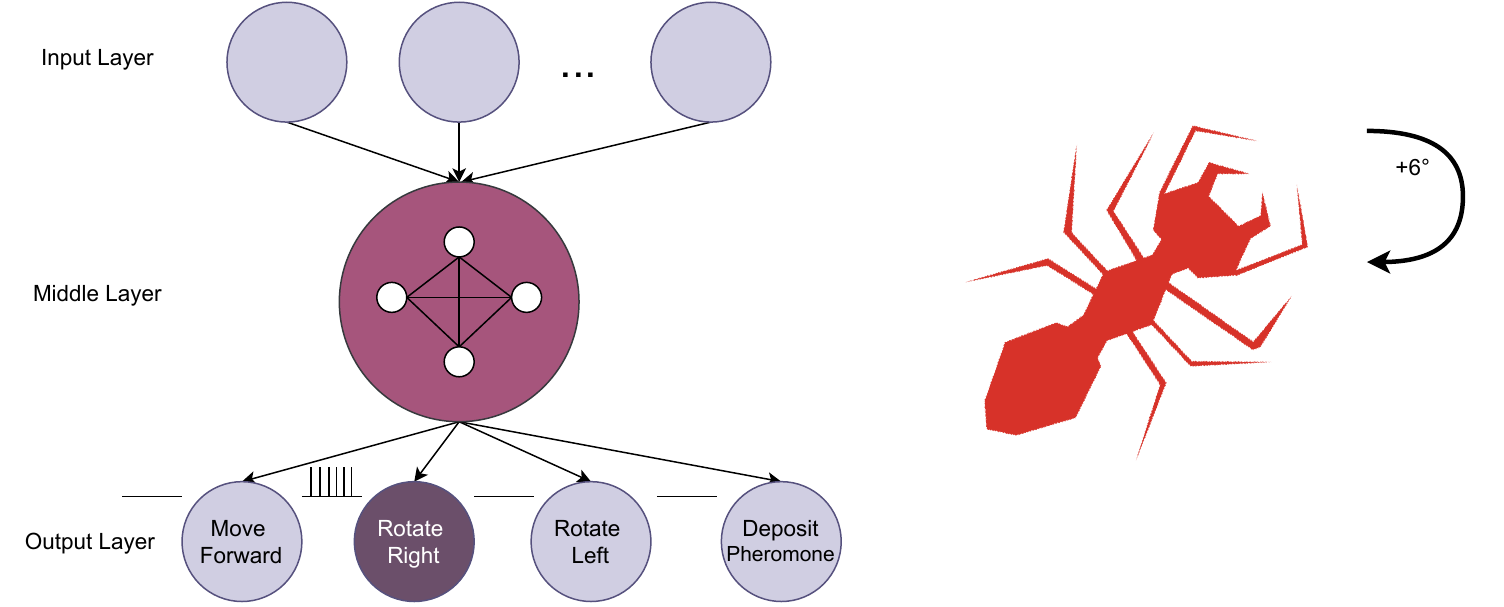}
    \caption{Actuation mechanism of the ant. The control network depicted on the left side, enables the ant to execute four distinct actions: advancing forward, rotating clockwise, rotating anticlockwise, and depositing a predetermined quantity of pheromones. To achieve this, each action is associated with its own dedicated neuron within the network. As illustrated on the right side, the ant turns 6 degrees clockwise in response to the activation of the neuron associated to the rotate-right action, which is shown on the left. 
    When a neuron exhibits firing activity, it triggers the corresponding actuator, enabling the agent to carry out the desired action. Consequently, the agent has the capacity to engage in forward motion, rotation, and pheromone deposition concurrently. Notably, the rotational movements occur in discrete steps of 6 degrees per activation.
    Simultaneously, the agent's forward motion progresses at a rate of 1 patch per activation, contributing to its ability to traverse the environment.
    However, it is important to note that exclusivity is observed only between the clockwise and anticlockwise rotation actions. In these instances, the neuron displaying the higher spiking activity exerts dominance and determines the direction of rotation that the agent will undertake, all while adhering to the 6-degree step increments.
 }
    \label{fig:ant_act}
\end{figure}

\hl{The output of the SNNs is determined by the firing rate of the output neurons during the network's simulation. The interpretation of the output vector relies on the correlation between the neurons and their respective actions, a concept that is depicted in } \autoref{fig:ant_act}.
The four output neurons are fully connected with the middle layer and control the movement (rotate left, rotate right, go forward) as well the activation to drop the pheromone.

\subsubsection{Fitness calculation}\label{subsubsec:fitness_calc}
The total fitness of the ant colony optimization is calculated by summing up the rewards and punishments received by each of its ants in the colony during the course of a simulation.
An ant receives a small positive reward for touching a food patch, and a large reward for returning with food to the nest. This induces the ants to quickly return to their nest, whenever they find food. A small punishment is given at every time step to encourage the colony to complete the task as quickly as possible. 
At the same time, the ants receive a small punishment for every selected action to avoid excessive reactions, such as depositing too much pheromone or exhibiting too much movement.

The simulation takes $T = 2000$ steps to finish; if the ants are able to collect all the food before the simulation ends they receive a reward as well. 
This reward is the result of the difference between the total simulation time and the time $T_{s}$ the ants require to collect the food multiplied with a scalar $\eta$.
The whole process can be formalized as:
\begin{equation}
 f_{i} = \sum_{t=1}^{T_{s}} \left(\sum_{j=1}^{J} \mathcal{N}^{(t)}_{i,j} + \mathcal{F}^{(t)}_{i,j} + \mathcal{C}^{(t)}_{i,j}   \right) + \eta \left(T-T_{s}\right),
 \label{eq:fitness_antcolony}
\end{equation}
where $t=1, \ldots, T_{s}$ is the simulation step with $T_{s}\leq T$ and $T$ the total simulation time, $\eta$ is a scalar to weight the speed of the ants collecting the food. 
Within the sum, $i$ is a specific individual, $j=1,\ldots, J$ is an ant and $J$ is the total number of ants in a colony. $\mathcal{N}$ is the positive reward value for coming back to the nest with food, $\mathcal{F}$ is a positive value for touching the food and $\mathcal{C}$ is the punishment cost. The behavioural reward and cost values are specified in the Appendix (Table~\ref{tab:fitness_cost}) and the other experimental parameters can be found in the Appendix (Table~\ref{tab:ga_ant_colony}) as well.

\section{Results}\label{sec:results}
In the following, we show the result of executing simulations of the foraging task using the L2L framework. 
See~\autoref{sec:appendix} for further details on the execution.
We explore and compare three models: 
\begin{enumerate}
    \item an SNN-driven model \hl{(SNN model 1)}, where each ant is controlled by identical copies of a common network,
    \item the same SNN-driven model but with the pheromone pathway deactivated (sensing and depositing of pheromone is disabled, \hl{SNN model 2}),
    \item a rule-driven model based (\hl{Rule based}) on the implementation in NetLogo by~\citet{wilensky1997netlogo}. In this model each ant follows a set of predefined simple rules.
\end{enumerate}

\subsection{Fitness and emergent usage of pheromone}\label{subsec:fitness_pheromone}
\begin{figure}
    \centering
    \includegraphics[width=0.7\textwidth]{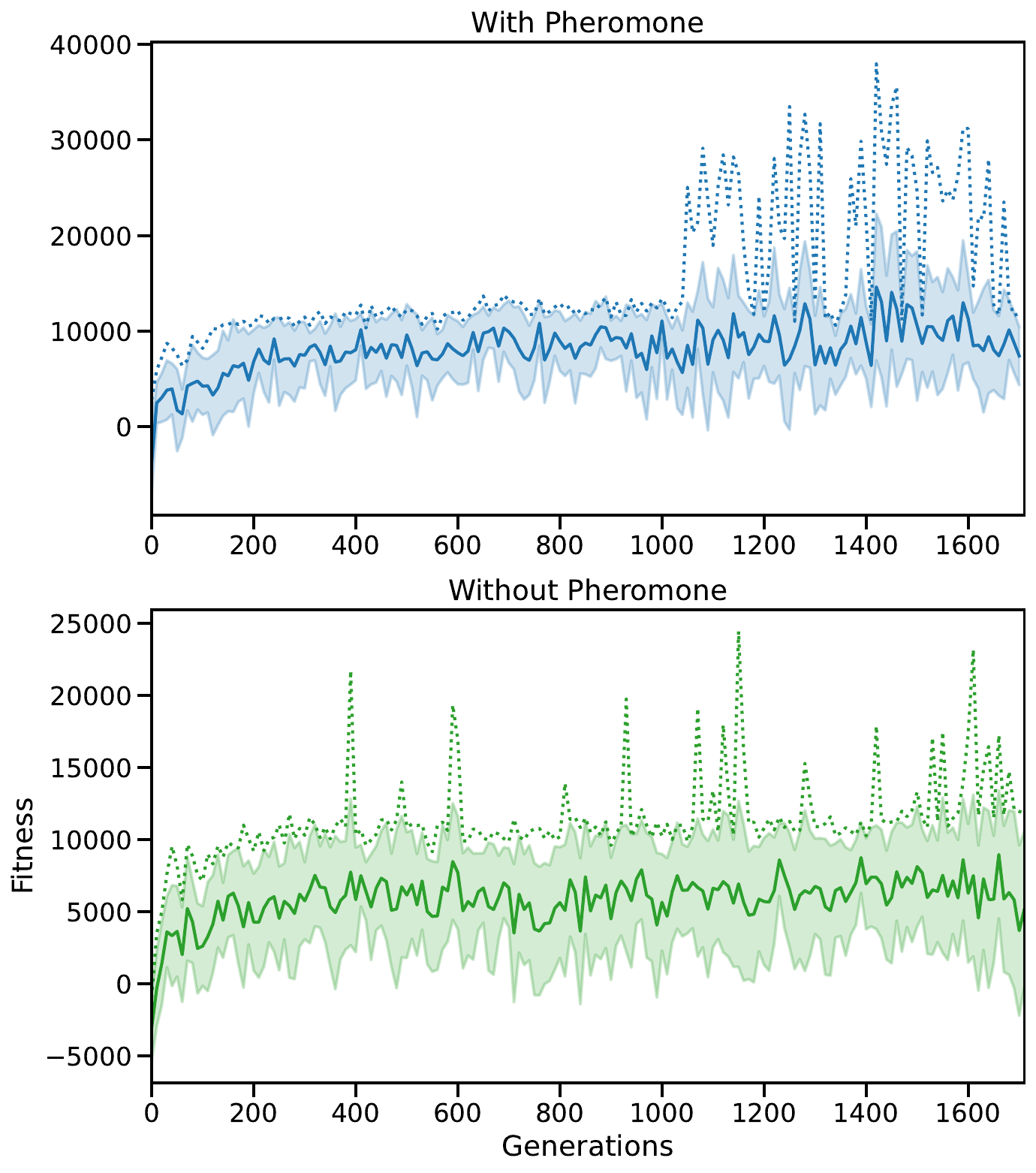}
        \caption{\label{fig:fitness_ant_colony_20ns} Fitness of the ant colony of two distinct simulation runs. In the first SNN-driven model the pheromone pathway is activated (top). However, the ants of the second SNN-driven model (bottom) cannot sense and deposit pheromones. While the blue data series depicts the evolution of the first model's fitness, the green data series illustrates it for the second model. The dotted data series shows the best individual found in every generation. The blue/green solid line indicates the mean fitness and the blue/green shaded area is the standard deviation over all individuals.}
\end{figure}

\begin{figure}
    \centering
    \includegraphics[width=0.7\textwidth]{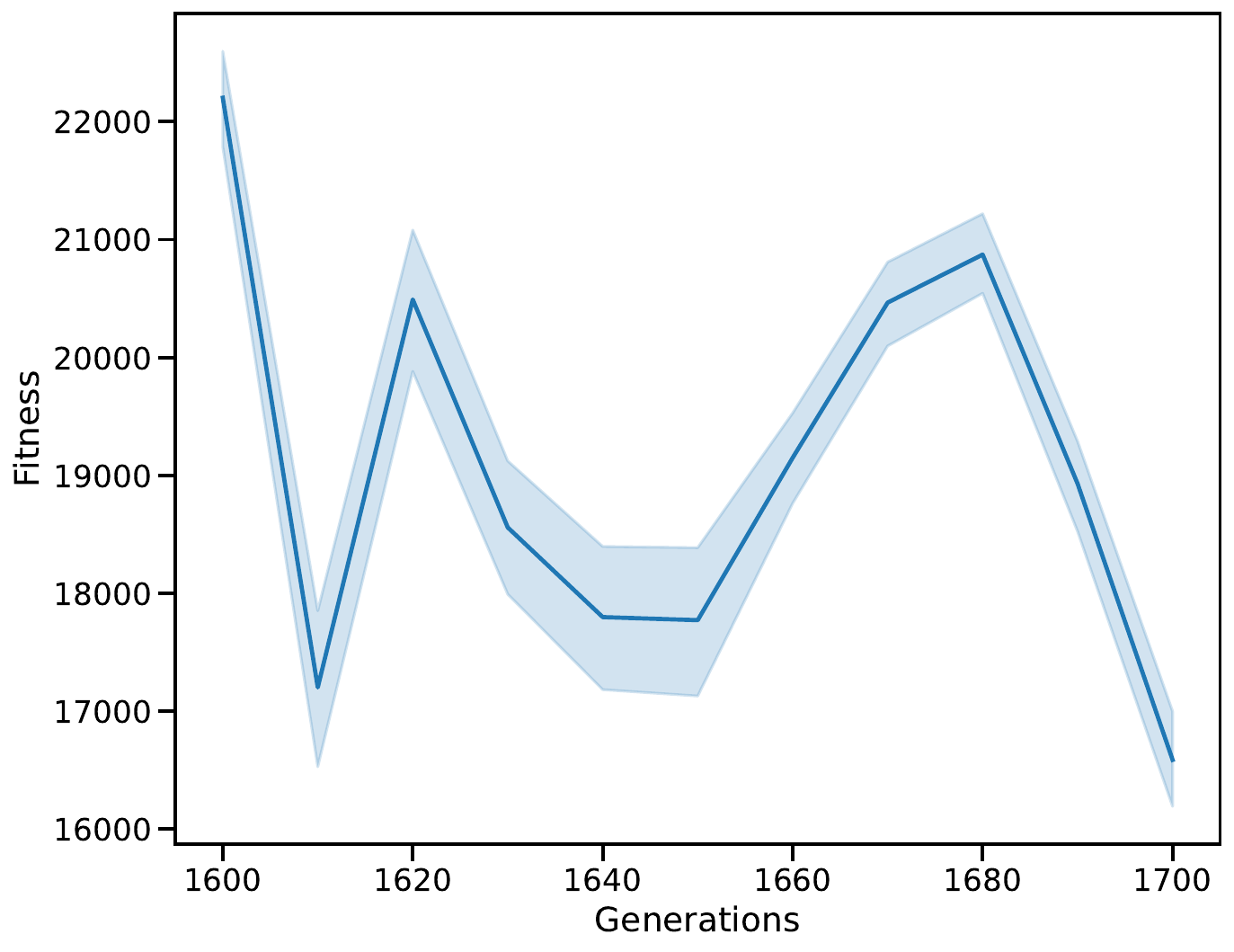}
    \caption{Performance of the best individual of the last hundred generations averaged over $100$ trials starting from generation $1600$. The shaded area depicts the confidence interval of $95\%$. The thick line shows the mean.}
    \label{fig:fitness_ci}
\end{figure}

\autoref{fig:fitness_ant_colony_20ns} shows the evolution of the mean and best fitness over $1700$ generations for the SNN-driven experiments mentioned above.
In the first SNN-driven model the pheromone pathway is enabled, i.e. the capability of the ants to deposit and sense the pheromone is active (\autoref{fig:fitness_ant_colony_20ns}, top).  
For this model, we observe that the mean fitness is increasing over the generations. 
After the $1000$th generation the mean converges and oscillates around a value of $8100$ with a standard deviation of $3970$. 
Specific individuals reach a fitness higher than $30000$ with the highest fitness at $38004$ in generation $1420$ (see green data series).
The second SNN-driven model has no active pheromone pathway, the ants are not able to perceive or deposit the pheromone (\autoref{fig:fitness_ant_colony_20ns}, bottom). 
Here, we observe an increasing mean, which converges after $700$ generations around a value of $5800$ with a standard deviation of $3950$. 
In comparison to the first model, specific individuals reach a high fitness in early generations (e.g. generation $390$ with a fitness of $21716$) and the highest fitness at $24450$ in generation $1150$. 
Furthermore, the colonies that are able to communicate via pheromones can be adapted via learning to achieve a higher fitness and performance than the colony without the ability to utilize pheromone. 

\autoref{fig:fitness_ci} depicts the performance of the best individual of the last hundred generations averaged over $100$ trials. Since the performance converges around the $1000$th generation we take a sample of the last hundred generations.
Each run is an independent simulation, where the position of the nest changes.
As observable in the plot, the best performing individual is at generation $1600$.
The observed oscillation of the fitness during the evolutionary optimization may arise from various factors, including the dynamic nature of the search space, genetic operators like crossover and mutation, and the algorithm's response to local optima. 
Stochasticity and population size also play a role. 
%
\begin{figure}[t]
    \centering
    \includegraphics[width=0.8\columnwidth]{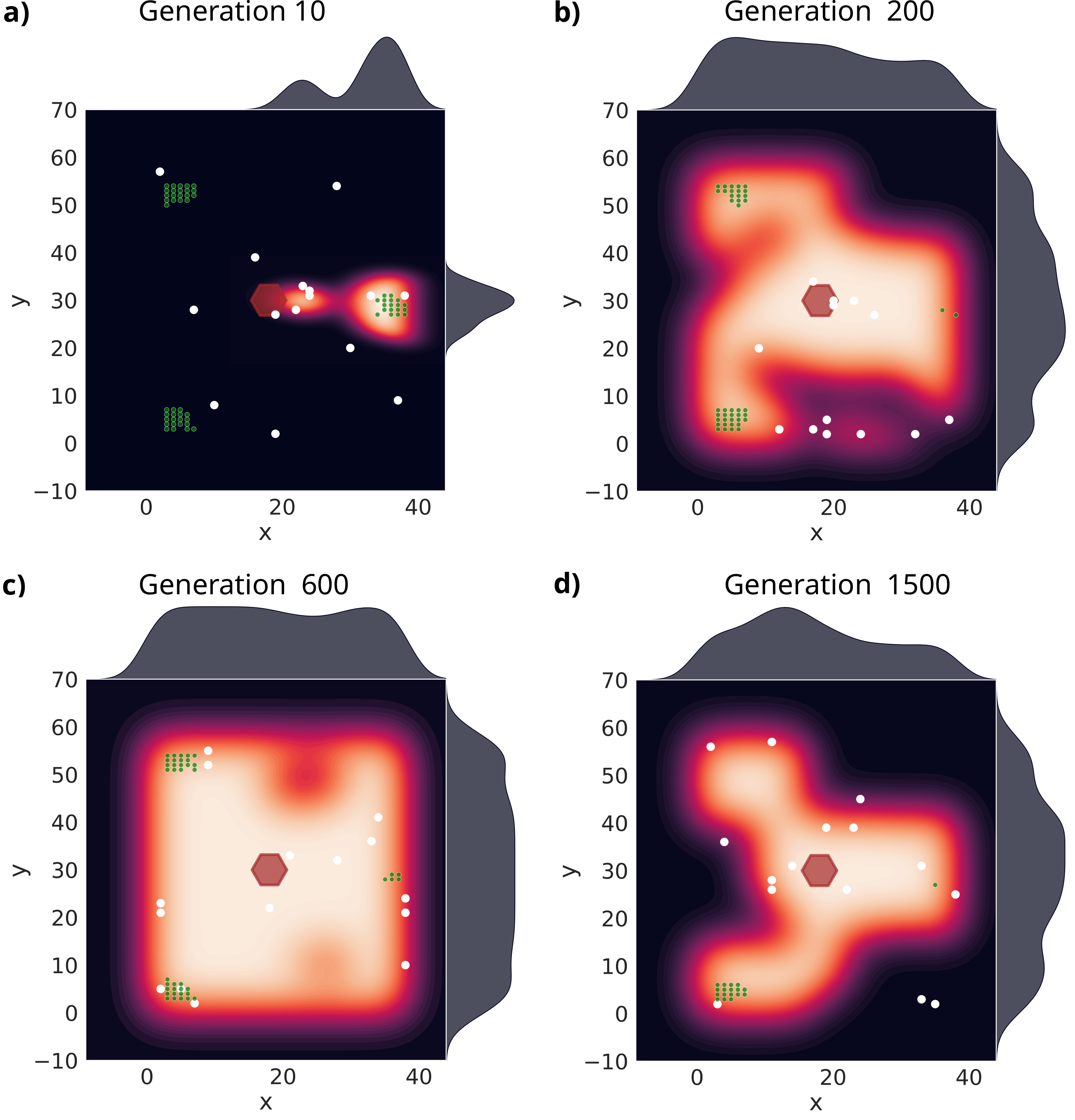}
    \caption{\label{fig:pheromone_concentration} Pheromone concentration heatmap over different generations. The brighter the color the higher the pheromone concentration. The marginal plots at the sides are density plots, indicating the amount of the concentration with regards to the $(x,y)$ position. The ants are colored in white, food patches in green. The brown hexagon in the middle is the nest.}
  \end{figure}
Beyond the evolution of the fitness, we examined the emergence of coordination via pheromones. 
The heatmap in \autoref{fig:pheromone_concentration} shows the pheromone concentration for an example individual over different generations.
Each simulation runs for $2000$ steps and at step $1900$ we take a snapshot of the best individual with the position of the ants, the pheromone and the food patches, as well as the pheromone concentration and the food amount.

In the early generations (\autoref{fig:pheromone_concentration}a) the concentration is low, while most of the food patches are uncollected.
The ants are 
exploring the environment \hl{in a random manner} and, from visual observation (as seen in \autoref{fig:pheromone_concentration}a), about half of the population is close to the nest as well as to the food pile on the right side ($x:35, \, y:30$). 
Small pheromone patches can already be observed around the nest and the closest food patch.
This is also visible in the marginal plots depicting the pheromone density. 
The peaks are at the position of the nest ($x:22, \, y:30$) and the right food pile. 
In generation $200$ (\autoref{fig:pheromone_concentration}b) it can be observed that there is a correlation between the pheromone concentration and the position of the food piles and the nest. 
\hl{Given the pheromone concentration and the position we hypothesize that the ants are utilizing the pheromone to keep track of food locations.}
The marginal plots have $4$ peaks, at the nest and at the left and right food piles (bottom left $x: 5, \, y: 5$, top left $x: 5, \, y: 52$, right $x:35, \, y:28$). 
In generation $600$ a higher pheromone usage is visible (\autoref{fig:pheromone_concentration}c) in comparison to generation $200$.
The concentration is nearly equally distributed over the environment, which is also visible in the density plots.
\hl{Since the pheromone depositing has a negative cost in the fitness function, continuous evolution of the network should optimize the pheromone usage for important information that is useful to accomplish the foraging task.}
\hl{This becomes visible in} \autoref{fig:pheromone_concentration}d \hl{ where the evolutionary pressure helps to transform the homogeneous pheromone distribution into well defined pathways. These paths connect the relevant components of the environment, i.e.~food location and the nest, such that foraging becomes faster and more efficient, leading to an overall higher fitness} (see \autoref{fig:fitness_ant_colony_20ns}, top).  

\subsection{Performance comparison}\label{subsec:performance_comparison}
\begin{figure}
\centering
    \includegraphics[width=0.75\textwidth]{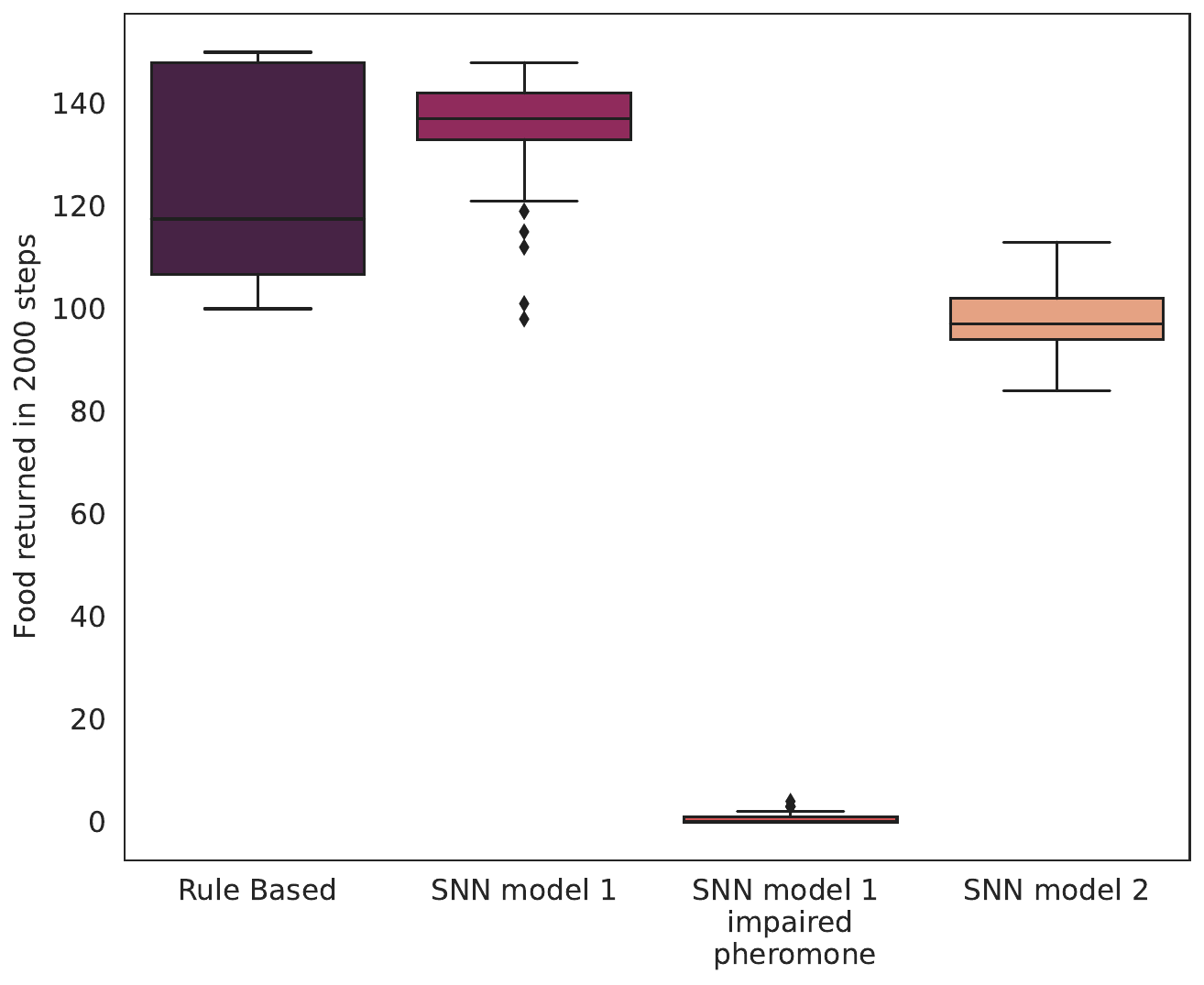}
    \caption{\label{fig:performance_comparison} Performance comparison of different ant colony models. The boxes depict the number of food patches collected in 2000 simulation steps in $100$ trials. Each box plot ranges from the first to the third quartile. The vertical bar in the box indicates the median and the whiskers depict the lowest or highest amount of collected food. The black diamonds show outliers. 
    In every trial the position of the nest is changed by setting a different seed.}
\end{figure}

To assess the performance of the ant colony, we first selected the best performing individual of the last generation after the optimization procedure. Secondly, we calculated over $100$ simulations (trials) the average amount of food foraged by the colony (and brought back to the nest) within the simulation time of $2000$ steps.
The trials have different seeds which determine the initial position of the ants in each trial.
There are three piles of food positioned in different areas of the simulated 2D-world as shown in \autoref{fig:pheromone_concentration}.
Each pile contains $50$ units of food making a total of $150$ units, which is the maximum score that a colony can reach in this experimental setup.

We tested the three models introduced in the beginning of \autoref{sec:results}.
Furthermore, we added a fourth experiment in which we impaired the pheromone pathway for the first SNN based model. 
The results are shown in \autoref{fig:performance_comparison} \hl{as box plots}.
The first box illustrates the foraging performance of a rule based system implemented in NetLogo~\citep{wilensky1997netlogo}, i.e.,~the ant foraging behaviour follows pre-defined action rules.
The self-organized behaviour of the ants emerges from the interactions of individual colony members.
In this setting, the colony reaches a \hl{median of $117.5$} foraged units of food with a standard deviation of $19.6$.
We define the rule based system as our baseline for comparison.
The second bar depicts the SNN based model 1 which reaches both a higher median score ($137.0$) and a lower standard deviation ($9.0$) than the rule-based system.
To measure the effect of the pheromone on the colony coordination, we disable the pheromone sensing in the first SNN model after evolving it. 
We observe that the foraging performance drops significantly and the ants are not able to collect the food (\hl{median:~$0.0$}, SD:~$0.9$) as shown in the third bar.
The fourth bar shows the performance of the second SNN based model. 
It reaches a \hl{median of $97.0$} with a standard deviation of $60.0$ and performs less efficient than the first SNN based model and the rule based one.

Notably, while we observe that SNN model 2 performs less effectively in food foraging compared to both SNN model 1 and the rule-based model, it still demonstrates the swarm's ability to explore the environment in search of food and return it to the nest.

\hl{
Several possibilities could explain this phenomenon:
1) Random chance. In a relatively small environment or with ants having a longer sensing range, ants might discover food sources by random chance, even without the guidance of pheromones. This could lead to effective foraging without the need for pheromones.
2) Alternative sensing mechanisms. SNN model 2 may have evolved synaptic pathways that exploit sensory information other than pheromones to locate and retrieve food. For example, ants might rely more on the food and wall sensors.
3) Parallel evolution. It's possible that the evolution process in SNN model 2 favored solutions that didn't critically rely on pheromones. These solutions could involve more robust and versatile strategies that didn't require pheromone communication.

The key question then arises: why doesn't SNN model 1 evolve to exploit the same information as SNN model 2 reducing its reliance on pheromones?
The answer may lie in the initial conditions and random variations inherent in evolutionary processes. SNN model 1 may have initially favored or stumbled upon solutions that heavily rely on pheromones due to the specific genetic and environmental conditions it encountered during the evolution process. It might not have explored or discovered alternative strategies or pathways as effectively as SNN model 2.

Overall, this situation underscores the complexity of evolutionary processes and the role of environmental factors in shaping the behavior of evolved agents in swarm systems. It also highlights the adaptability of spiking neural networks in finding effective solutions in different scenarios.}

\hl{Based on these results, we observe our evolved model with a self-coordination system is able to forage for food in a comparable amount of time as the rule-based system but faster than the second SNN model which is evolved solely using the visual sensory pathways to navigate and forage for food.}
The ants are able to learn to communicate via pheromones, which increases their performance when foraging for food, despite the fact that there is no hardwired mapping that defines the usage of pheromones for communication purposes.
The communication emerges within the ant colony over the generations as a swarm strategy.

\subsection{Input-output correlation analysis}\label{subsec:network_analysis}
\begin{figure}[ht]
    \centering
    \includegraphics[width=1.0\columnwidth]{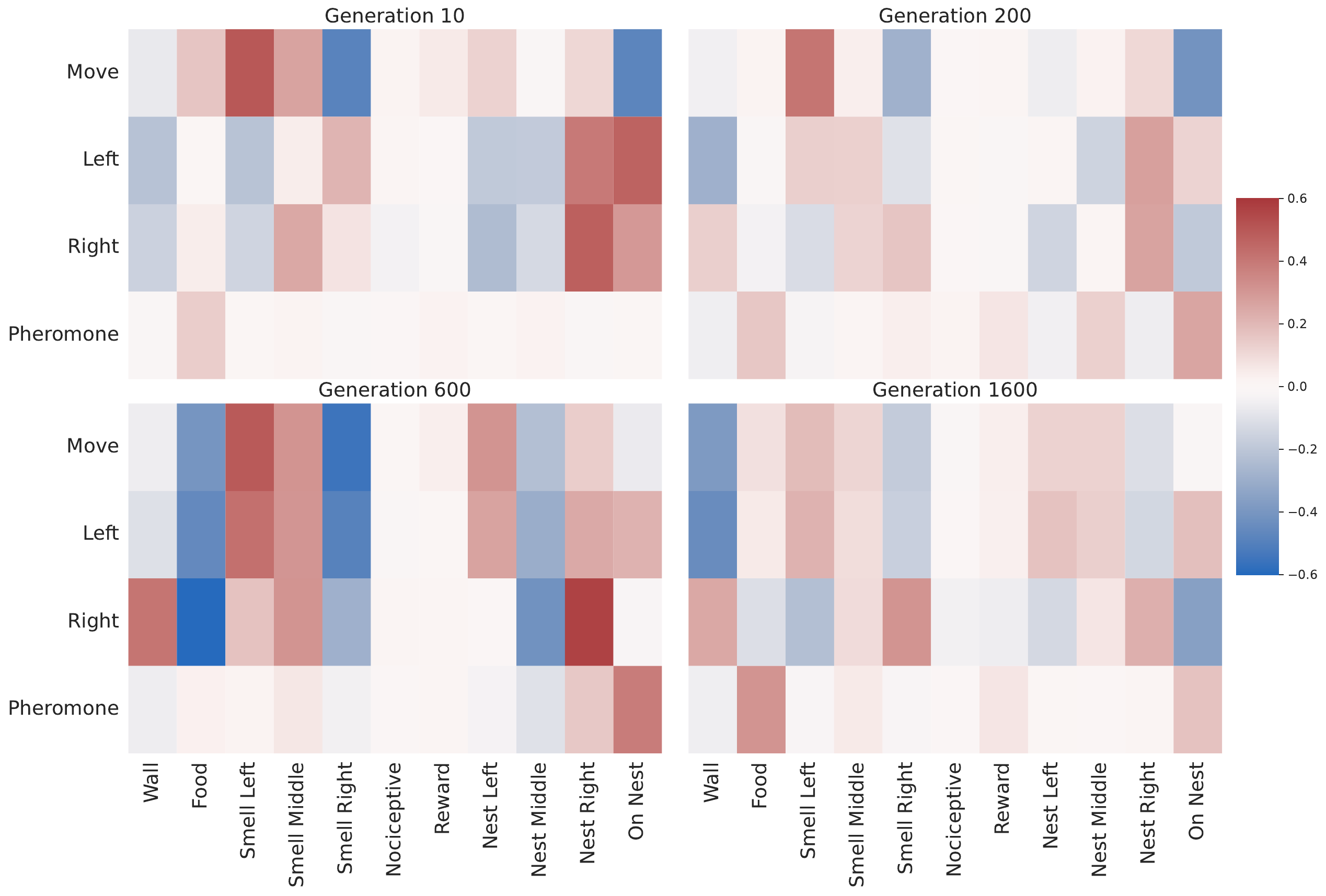}
    \caption{\label{fig:correlation_map} Correlation heatmap between the network input and output spike trains. The mean Pearson product-moment correlation coefficients between the input and output spike trains of all ants from the best individual are calculated.}
\end{figure}

In order to better understand how the ant behaviour changes over the generations, we correlate all input spike trains with all output spike trains for all ants of the best individual for specific generations.
The results of this correlation analysis are limited and provide just an initial overview of the relationships between input and output in the network. 
A more complete assessment of the correlations in a recurrent network with dynamic inputs requires a complex analysis which considers not only snapshots of inputs and outputs but also the internal state of the model and its transitions through time. 
This type of analysis is out of scope for this work and will be the topic of further publications.

First, we create a histogram to bin the input and output spike trains using the Electrophysiology Analysis Toolkit \citep[Elephant\footnote{Version 0.11.1};][]{elephant18}.
As a second step, we obtain the mean Pearson product-moment correlation coefficients.
\autoref{fig:correlation_map} depicts the correlation heatmap for the input (x-axis) and the output (y-axis) activity (see also \autoref{fig:colony_net} for the network description) for four different generations.
While a red color indicates a high correlation, a blue color denotes a negative correlation.
The nomenclature follows the naming scheme in \autoref{fig:colony_net}, i.e.~for the output activity \textit{Left} is rotate left, \textit{Move} means move forward etc. and similarly for the input activity.

\hl{As optimization of the network takes places we observe that a correlation between smelling and moving in the same direction emerges.
Given this observation we hypothesise that the ants develop a mechanism to attract other ants to specific locations of interest using pheromones.}
Across all generations, \textit{Smell Middle} is positively correlated with \textit{Move}, 
with a value of $0.265$ in generation $10$ and a reduced value of $0.116$ in generation $1600$. 
One can also see that e.g.~\textit{Smell Right} is negatively correlated with \textit{Move}, it has a value of $-0.487$ in generation $10$, but at generation $1600$ the strength of this anti-correlation is reduced to $-0.182$.
Interestingly, \textit{Smell Left} is negatively correlated with \textit{Left} in generation $10$ ($-0.213$) but shows a positive correlation in later generations ($0.219$ in generation $1600$). 
A similar observation can be made for \textit{Smell Right} and \textit{Right}. 

We observe in early generations a negative correlation between \textit{Move} and \textit{On Nest} ($-0.478$ in generation $10$), while in later generations there is a trend towards becoming uncorrelated ($-0.0017$ in generation $1600$).
\textit{Nest Middle} and \textit{Left}  are anti-correlated in early generations ($-0.187$ in generation $10$) and positively correlated in later generations ($0.135$ generation $1600$).
The relationship between \textit{Food} and \textit{Pheromone} indicates a slight positive correlation in early generations ($0.136$ in generation $10$) but increases in later generations ($0.313$ in generation $1600$).
\hl{The observation that ants tend to deposit pheromones when they encounter a green patch suggests that they likely use pheromones to attract other ants to interesting locations, such as food sources.
This relationship between visual stimuli and pheromone sensing appears to have developed and improved over the course of evolution, eventually leading to ants relying more on pheromone signals than visual cues.
To test this hypothesis we disabled the pheromone sensing from good individuals which were originally evolved with both visual and pheromone sensors. The results of this experiment confirmed that the presence of pheromone sensing became essential for these ants to successfully accomplish their foraging task} (see \autoref{subsec:fitness_pheromone}). 
The fitness function punishes an extensive pheromone drop which  leads to a rather restrained usage.
The input-output relationship is changing continuously across the generations due to the evolutionary algorithm, i.e.~certain input-output relationships get enhanced or weakened. 
Some of these mappings align with observations in nature, e.g.~\textit{Food} and \textit{Pheromone} have an increasing correlation value across the generations, which reflects the positive reinforcement when the ant perceives food. 
This is also in agreement with the attractor effect of the pheromone and links to the correlations between smelling pheromones and ant movement, serving as guiding signals to take the ants back to the food patches.  

\subsection{Correlation analysis over all generations}\label{subsec:corr_all_gens}
\begin{figure}[t]
  \centering
  \includegraphics[width=1.0\columnwidth]{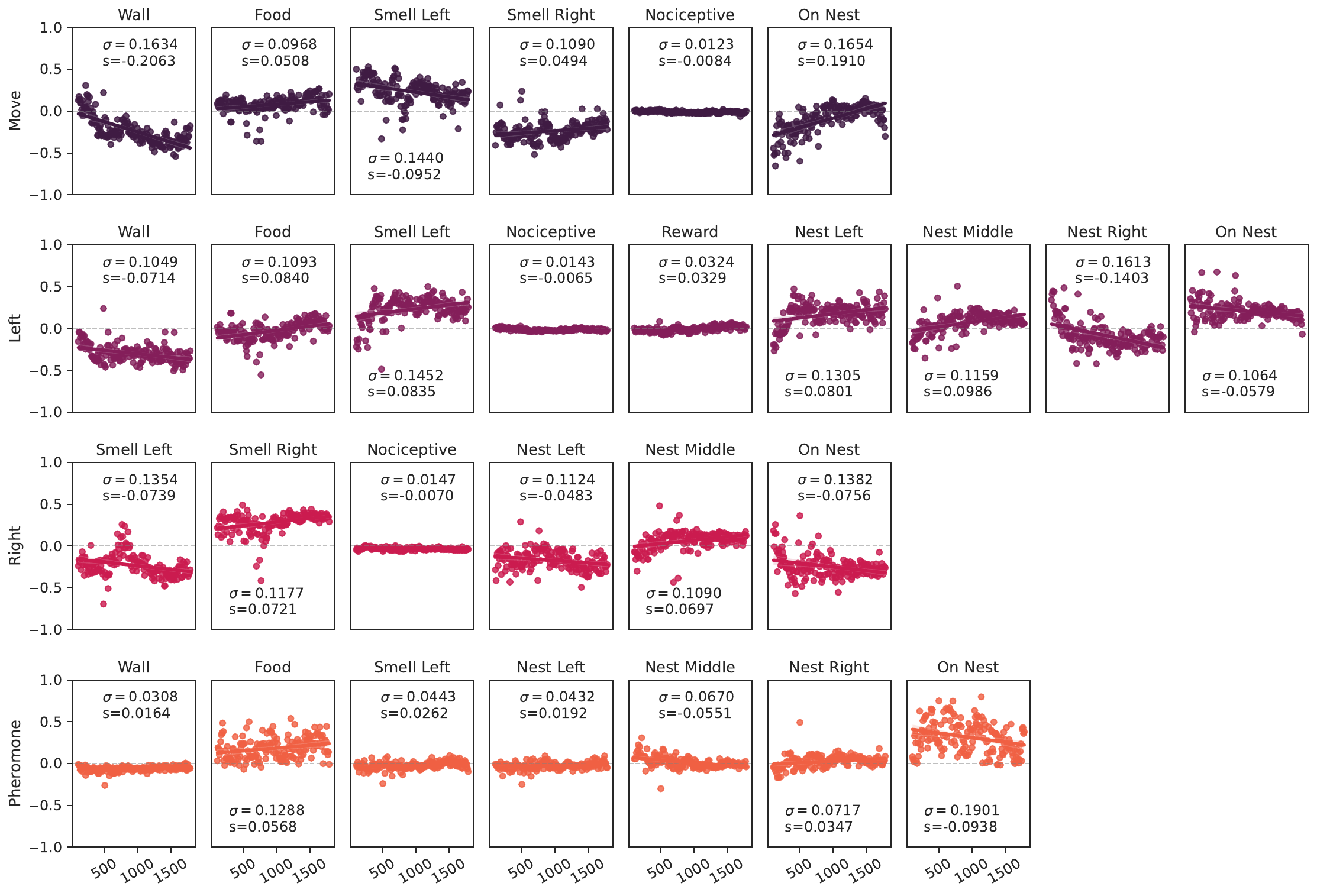}
  \caption{\label{fig:correlation_regression}Correlation coefficients of input and output spike trains for the first ant of the best individual every ten generations. A linear regression is calculated for the data (solid lines).  
  A low p-value rejects the null-hypothesis that the slope of the curve is zero. 
  s is the slope of the curve and $\sigma$ the standard deviation of the points.}
\end{figure}

Although the pheromone heatmap provides insight into ant behaviour for specific generations, we wanted to see how the behaviour evolves over all generations.
In order to compare different input-to-output correlations and validate our results obtained from the heatmap, we calculate the correlation coefficients between input and output spike trains for the first ant of the best individual every tenth generation.
We then fit a linear regression on the coefficients, which returns a p-value and the slope s; a low p-value rejects the null-hypothesis that the slope of the curve is zero, i.e.~there is not a correlation.
Additionally, we calculate the standard deviation $\sigma$ of all points. 

The visualization in \autoref{fig:correlation_regression} enables us to determine trends in the ant behavior over generations. 
\hl{Note, that only plots with a p-value below $0.005$ are displayed.}
\hl{We hypothesise that the ants progressively learn to avoid the red walls and turn away when approaching them. 
In contrast, they seem to move towards the food patches and the nest based on sensing the pheromones or on visual stimuli.}
For example, we observe that \textit{Wall} is initially positively correlated to  \textit{Move} but there is a trend towards anti-correlation over the generations. 
While \textit{Food} and \textit{Move} show an increasing trend, \textit{Wall} and \textit{Move} become increasingly negatively correlated over the generations. 
Likewise, \textit{Smell Right} shows a trend towards an increasing positive correlation with \textit{Right} and a negative correlation with \textit{Left}. 
\textit{Smell Left} is initially positively correlated to \textit{Move}, but this correlation decreases for later generations.
In contrast, the correlation between \textit{Smell Left} and \textit{Nest Left} increases, while it decreases for \textit{Nest Right}. 
\hl{It appears that the ants developed a perception for the nest and turn towards the corresponding direction.}
The sensor for \textit{Smell Middle} shows an increasing trend towards positive correlation with \textit{Move}.
\textit{Nest Middle} has an increasing positive correlation with \textit{Left}, while \textit{Nest Right} has an increasing negative correlation with \textit{Left} as the colony evolves.
\hl{We further hypothesize that the ants have a tendency to move after they visited the nest in order to explore the environment or forage for food. 
At the end of the evolution there is a positive correlation between \textit{Move} and \textit{On Nest} and a negative correlation with \textit{Left} or \textit{Right} and \textit{On Nest}.}
Pheromone is initially dropped in the vicinity of the nest. 
However, the pheromone output activity, when the ant is in the middle of the nest, decreases over the generations. 
In contrast, \textit{Food} and \textit{Pheromone} show an increasing positive trend.

While observing the performance of the ant colony especially at the end of the evolutionary process, pheromone trails emerge and form attractors with higher pheromone concentration. 
This results from a combination of the evolutionary pressure of the fitness metric towards the pattern: \texttt{explore, get food, return to nest}, as well as the optimized actions which includes sensing the pheromone and exploiting the pheromone dynamics (evaporation and diffusion).
This resembles the actual behaviour of ant colonies observed in nature.  

\section{Discussion and Conclusion}\label{sec:discussion}
In this work we discussed an implementation of a virtual environment where the behaviour of swarms of agents controlled by SNNs could be optimized using evolutionary algorithms. 
The implementation combines NEST simulations for the SNNs, an environment described with NetLogo and an optimization with genetic algorithms using the L2L framework. 
This implementation is open source (see~\autoref{sec:appendix}) and can be used or extended to explore a variety of scientific questions regarding swarm optimization as well as emergent communication and organization. 
In particular, it could be of high interest to study the coordinated emergent behaviour and communication between agents in neurorobotics and cobotics applications, where it is essential to ensure safety, optimal resource usage, and adaptation to dynamic environments. 

We demonstrated that we could use genetic algorithms to successfully evolve spiking neural networks to control agents (here, ants) to solve a multi-agent food foraging task.
Our approach shows emerging self coordination via pheromone as a result of the network optimization, without defining action rules or probabilistic state models. 
This work serves as a proof of concept to illustrate multi-agent interactions with emergent communication and self-coordination utilizing SNNs as underlying architectures.
We provide a short summary of our methods in the following:
\begin{itemize}
    \item Our SNN architecture encodes the inputs from the environment and steers the behaviour of the ant when foraging for food. 
    \item Based on the fitness function, which evaluates the foraging performance, we utilize a genetic algorithm to optimize the synaptic weights and delays. This allows the network to achieve an efficient foraging behaviour.
    \item Our analysis investigates the emergent relationship established by the SNN between the inputs and outputs to trigger certain actions during the evolutionary process.
\end{itemize}
%

By applying the methods mentioned above, we observed an emergence of stigmergy, i.e.~the communication and collaboration via pheromones.
In our evolved network we proved that communication became essential for the efficient performance of the ants during the foraging task. 
Impairing the pheromone sensing capabilities of the ants drastically hindered the performance of the collective.
This highlights the importance of stigmergic communication, which emerged through the optimization process, for the effective collaboration between the agents.
In our experiments we showed that agents using pheromones were able to quickly forage for food even when the the position of the nest was changed.
It is important to note that the release of pheromone does not obey any pre-defined rule or manual synaptic pre-configuration in the network.
Instead, the stigmergy is an emergent behaviour that is triggered under certain conditions established during the evolutionary process. 

Furthermore, we would like to highlight that the sensing of the pheromone is not associated with any kind of pre-defined behavioural response in the network.
There is no hard-coded synaptic communication that explicitly maps the sensory input of pheromone with a certain action. 
Instead, the observed behaviour is completely based on emergent communication resulting from the collective evolution of the colony.
What we observe is that the optimization operates directly on the structure (synapse weights) of the brain networks of the ants, but there is also a higher level optimization in order to create a functional network created by the collaborative behaviour of the agents. 
This behaviour is achieved by the development of communication strategies between the agents using the pheromone signals. 
In essence, the ants form a new network where pheromone signals are used to coordinate the movement of the agents as a whole to efficiently solve the task.
We can see this emergent functional effect as the correlations and anti-correlations between directed movement and the sensing of pheromone increase with each generation (see \autoref{sec:results}).
Based on \autoref{fig:correlation_map} it is possible to arrive at the conclusion that the pheromone dropping is uncorrelated to actions and sensing of the environment. 
However, optimized movement based on pheromone signals makes it possible to have higher concentrations of pheromone around the areas of interest and thus enhance the ability of the swarm to collectively solve the task.
We also confirmed the emergence of this functional network when we disabled the pheromone sensor and the colony was not able to efficiently solve the task anymore (see \autoref{subsec:performance_comparison}).

Interestingly, the emergent communication via a simulated chemical signal resembles the use of positive (attractor) pheromones observed in real ant colonies.
Although attractive trails of pheromone are more commonly found in nature, it has been observed that negative (repelling) pheromones are also used in certain species of ants ~\citep{NegPheromone2005}. This negative pheromone also known as a ``no-entry" signal is used by the ants to mark unrewarding paths to prevent other cohorts from using them. 
It opens the question of whether further experimentation simulating different evolutionary trajectories could lead to the emergence of negative repelling pheromones among other sophisticated coordination mechanisms in self-organized simulated swarms. 
Another related question would be to study which environmental and evolutionary requirements are necessary for different signals to emerge within the swarm's communication repertoire.
The usage of the L2L framework allows the modification of the outer loop optimization algorithm, opening the possibility to explore the impact of different evolutionary optimization features in the resulting communication patterns.

At the beginning of this work we formulated two research questions. 
The results presented in the previous sections answer these. 
First, the usage of pheromone as a mean of communication and self-organized coordination between agents can emerge in the absence of predefined rules. 
Neither the mechanism to trigger the release of pheromone nor the mechanism of interpreting these signals was engineered as part of the behaviour of the agents. 
Through this we can infer that the simulated embedded physiological properties of the agents and the characteristics of the environment are exploited by the optimization algorithm.
To answer the second research question, the results in \autoref{subsec:performance_comparison} illustrate the difference in the performance of the colony in terms of foraging efficiency.

\hl{The evolved SNN-based colony with stigmergic communication achieved a performance comparable to the multi-agent rule-based colony.}

We would also like to address some of the main shortcomings of our work. 
First, the genetic algorithm requires several hundreds of generations until it provides a parameter setting which leads to a suitable task performance. 
We want to investigate other optimization techniques which may decrease the number of generations and result in a similar performance. 
Second, in this work we use a manually defined network architecture for all individuals. 
This limits the optimization process and the potential functional capabilities of the ants. 
Future work optimizing the network architecture could help find new strategies of the ants to use their physiology to efficiently perform the foraging task.
Third, even if we did not define any rule about the relationships between input and output actions of the ants, the fitness function used to optimize the network is static and manually defined. 
This imposes evolutionary pressure on the ants towards specific solutions. Exploring other combinations of the fitness metrics will help us, in future experiments, understand better the strength of such pressure in the emergence of specific behaviour and solutions by the colony.
Finally, as discussed in the results section, our analysis of the correlations between input and output is also limited and further work needs to be done to fully understand the relationships within the ant networks and the colony as a whole through time. 

\section{Future Work}\label{sec:future_work}
As future work we want to analyze further the emergence of the communication features in the functional network created by the collective of ants and how this evolves through time guided by the genetic algorithm.
In particular, we want to measure the effect of pheromone concentration and evaporation rate in the communication strategy. 
A way to do this would be to represent the whole colony as functional network where connectivity is defined by spatial coincidence in the environment between one or more ants, signals are represented by pheromone concentration and delays by the pheromone evaporation rate.
We also want to explore the usage of plasticity within the ant networks and optimize the learning capabilities of the ants in the inner loop of L2L and see if this capability could further enhance the performance of the swarm.
For example, we want to learn the foraging task utilizing synaptic plasticity in the inner loop and optimize the model's hyper-parameters in the outer loop. 
In particular, the emergence of connectivity motifs, as emerging from the usage of STDP in \citet{ocker2015self}, could be further studied and correlated to specific communication functions in the colony.

During pilot experiments (not shown in this manuscript) we observed that the emergence of stigmergy could be influenced by the number of neurons in the network of each agent. 
As future work we want to explore the minimal structural requirements (number of neurons, topology, types of synapses, etc) for the SNNs to enable the emergence of collective communication and its effect on functional efficiency of the swarm as a whole.

The emergent foraging behaviour that has been described in this work resembles to some extent the self-organisation observed in social insects. 
Self-organisation is a process, where high order structure emerges \citep{kauffman1993origins}. 
The emergent behaviour allows the swarm to collectively adapt to changing environments where the colonies become robust systems even if some individuals fail to perform their tasks. 
Robustness and adaptability are desired characteristics in the application of robotics,
specially for the deployment in changing environments with harsh or extreme conditions where single agents are more prone to failure (e.g. space exploration, undersea navigation, disaster areas~\citep{staudinger2018swarm,dorigo2020reflections}).  

\newpage
\section*{Appendix}\label{sec:appendix}
\begin{table}[ht]
\begin{center}
\begin{tabular}{|l|l|}
\hline
Parameter     & Setting \\ \hline \hline
Resting potential & $-70.0$mV \\ \hline
Spike threshold & $-55$mV \\ \hline
Reset potential & $-70.0$mV \\ \hline
Refractory duration & $1.0$ms\\ \hline
Default amplitude & $2000$pA \\ \hline
\end{tabular}
\end{center}
\caption{\label{tab:lif_parameters} Parameters of the leaky-integrate-and-fire neuron model.}
\end{table}
\begin{table}[h]
\begin{center}
\begin{tabular}{|l|l|}
\hline
Behavior     & Cost \\ \hline \hline
Dropping Pheromone & -0.05 \\ \hline
Rotation & -0.02 \\ \hline
Movement & -0.25 \\ \hline
Return nest & 220\\ \hline
Touch food & 1.5 \\ \hline
$\eta$ & 30.0 \\ \hline
\end{tabular}
\end{center}
\caption{\label{tab:fitness_cost}Cost values corresponding to the behaviour of an ant utilized in the fitness function. Every movement, rotation and pheromone drop induces a small punishment while returning to the nest with food and touching the food is rewarded.}\label{tab:cost_value}
\end{table}

\subsection{L2L:~Technical description of the inner loop}\label{subsec:l2l_details}
The simulations of our models run in the inner loop.
After every simulation the performance of the agents is evaluated.
To execute the simulation in L2L three functions have to be defined.
\begin{itemize}
    \item \texttt{create\_individual()}: One individual is a set of parameters which is to be optimized. In our case we have connections weights and delays. \hl{This function is called only once in the initialization phase of the optimization process.}
    \item \texttt{simulate()}: In this function the model is invoked and a simulation starts. Additionally, the simulation workflow is steered here.
    \item $\texttt{bounding\_func()}$: It is possible that in the optimization process parameter ranges can be exceeded (e.g. negative delays). In our setting, this function clips the parameters to defined ranges.
\end{itemize}

First, the optimizee instantiates an individual within the \texttt{create\_individual()} function
Here, weights and delays are created.
Weights are uniformly distributed in the range of $[-20, 20]$ and delays are in $[1, \ldots, 7)_{\mathbb{N^{+}}}$.
A min-max normalization can be applied on the parameters.
Before the simulation is started the parameters are saved into a csv file and later read in by the simulation when creating the network.
To run the simulation we use the Python subprocess routine in the \texttt{simulate()} function to invoke the headless mode of NetLogo which then starts the simulation and returns a fitness value after the simulation ends. The specific fitness parameters are defined in Table~\ref{tab:fitness_cost}.
The optimizee waits until the simulation is finished and collects the results, which are stored in the same aforementioned csv file.
Furthermore, to restrict the parameters to not exceed certain ranges we restrain them using the $\texttt{bounding\_func()}$ function.
Weights are clipped to the range of $[-20, 20]$ and delays to $[1, 5]$.

It is important to note that one inner loop has several individuals, i.e.,~each simulation of one individual is executed in parallel on the high performance computing system (HPC).
We conducted our experiments on the JUSUF HPC\footnote{\url{https://apps.fz-juelich.de/jsc/hps/jusuf/}} and on a workstation with an AMD Ryzen Threadripper CPU (32 cores), 64GB RAM and Ubuntu OS 21.04.

\subsection{L2L:~Outer loop optimization}
\begin{table}[h]
\begin{center}
   \begin{tabular}{|l|l|}
   \hline
   Parameter & Value \\ \hline \hline
   Cross-over probability & 0.7     \\ \hline
   Blend Cross-over $\alpha$ & 1.0 \\ \hline
   Mutation probability & 0.5 \\ \hline
   Probability of mutation of each element in individual & 0.08 \\ \hline
   Tournament size $k$ & 8 \\ \hline
   Standard deviation for the Gaussian addition & 0.5 \\ \hline
   Hall of Fame & 20 \\ \hline
   \end{tabular}
\end{center}
\caption{Genetic algorithm parameters}\label{tab:ga_ant_colony}
\end{table}
The optimization process which is started after the simulation is run and the optimizee sends the parameters to the outer loop.
We optimize utilizing the genetic algorithm (GA).
The GA also has its own hyper-parameters, which are initialized in the beginning.
In Table~\ref{tab:ga_ant_colony} we provide the hyper-parameters.

A high cross-over probability ensures a higher cross-over rate and favors a recombination of values (i.e., the chromosomes) between parent and child individuals.
The cross-over is applying the blend cross-over method which picks values between the parent chromosomes, but also allows to set a range ($\alpha$) outside the parent values.
In the tournament selection, the best $k$ individuals are chosen for the cross-over.
Best individuals in terms of fitness are stored in the Hall of Fame (HoF).
The mutation is responsible for keeping a diversity between the individuals by adding Gaussian noise to the individuals which go into the cross-over step.
In the mutation step every value of the individual can be perturbed if a randomly drawn probability is below a given threshold.
Higher values ensure a bigger variety in the chromosomes and enable a wider exploration for better performing individuals.

\section*{Conflict of Interest Statement}
The authors declare that the research was conducted in the absence of any commercial or financial relationships that could be construed as a potential conflict of interest.

\section*{Author Contributions}
All authors conceived of the project.
CR and AY worked on the design of the project.
CR, AY and APM worked on the implementation.
AY and CR produced the results reported in the manuscript.
All authors reviewed, contributed and approved the final version of the manuscript

\section*{Funding}
The research leading to these results has received funding from the European
Union’s Horizon 2020 Framework Programme for Research and Innovation
under the Specific Grant Agreement No.  945539 (Human Brain Project SGA3). This research has also been partially funded by the Helmholtz Association through the Helmholtz Portfolio Theme "Supercomputing and Modeling for the Human Brain". Open Access publication funded by the Deutsche Forschungsgemeinschaft (DFG, German Research Foundation) - 491111487.

\section*{Acknowledgments}
CR and AY would like to thank Chiara Segala for her input and feedback regarding the correlation analysis.
We would like to thank Wouter Klijn and Michael Herty for their ideas, support and useful input.
We acknowledge the use of Fenix Infrastructure resources, which are partially funded from the European Union's Horizon 2020 research and innovation programme through the ICEI project under the grant agreement No. 800858.
The authors gratefully acknowledge the Gauss Centre for Supercomputing e.V. (www.gauss-centre.eu) for funding this project by providing computing time on the GCS Supercomputer JUWELS at Jülich Supercomputing Centre (JSC).

\bibliographystyle{unsrtnat}
\bibliography{main.bib}

\begin{thebibliography}{35}
\providecommand{\natexlab}[1]{#1}
\providecommand{\url}[1]{\texttt{#1}}
\expandafter\ifx\csname urlstyle\endcsname\relax
  \providecommand{\doi}[1]{doi: #1}\else
  \providecommand{\doi}{doi: \begingroup \urlstyle{rm}\Url}\fi

\bibitem[Wilson et~al.(1967)Wilson, Carpenter, and Brown~Jr]{wilson1967first}
Edward~O Wilson, Frank~M Carpenter, and William~L Brown~Jr.
\newblock The first mesozoic ants.
\newblock \emph{Science}, 157\penalty0 (3792):\penalty0 1038--1040, 1967.

\bibitem[Smith et~al.(2008)Smith, Toth, Suarez, and Robinson]{smith2008genetic}
Chris~R Smith, Amy~L Toth, Andrew~V Suarez, and Gene~E Robinson.
\newblock Genetic and genomic analyses of the division of labour in insect
  societies.
\newblock \emph{Nature Reviews Genetics}, 9\penalty0 (10):\penalty0 735--748,
  2008.

\bibitem[Wilson and Nowak(2014)]{wilson2014natural}
Edward~O Wilson and Martin~A Nowak.
\newblock Natural selection drives the evolution of ant life cycles.
\newblock \emph{Proceedings of the National Academy of Sciences}, 111\penalty0
  (35):\penalty0 12585--12590, 2014.

\bibitem[Boudinot et~al.(2022)Boudinot, Richter, Katzke, Chaul, Keller,
  Economo, Beutel, and Yamamoto]{boudinot2022evidence}
Brendon~E Boudinot, Adrian Richter, Julian Katzke, J{\'u}lio~CM Chaul,
  Roberto~A Keller, Evan~P Economo, Rolf~Georg Beutel, and Sh{\^u}hei Yamamoto.
\newblock Evidence for the evolution of eusociality in stem ants and a
  systematic revision of† gerontoformica (hymenoptera: Formicidae).
\newblock \emph{Zoological Journal of the Linnean Society}, 195\penalty0
  (4):\penalty0 1355--1389, 2022.

\bibitem[Vittori et~al.(2004)Vittori, Gautrais, Ara{\'u}jo, Fourcassi{\'e}, and
  Theraulaz]{vittori2004modeling}
Karla Vittori, Jacques Gautrais, Aluizio~FR Ara{\'u}jo, Vincent Fourcassi{\'e},
  and Guy Theraulaz.
\newblock Modeling ant behavior under a variable environment.
\newblock In \emph{International Workshop on Ant Colony Optimization and Swarm
  Intelligence}, pages 190--201. Springer, 2004.

\bibitem[Bandeira~de Melo and Ara{\'u}jo(2008)]{bandeira2008modeling}
Elton~Bernardo Bandeira~de Melo and Alu{\'\i}zio Fausto~Ribeiro Ara{\'u}jo.
\newblock Modeling ant colony foraging in dynamic and confined environment.
\newblock In \emph{Proceedings of the 10th annual conference on Genetic and
  evolutionary computation}, pages 169--176, 2008.

\bibitem[Hecker and Moses(2015)]{hecker2015beyond}
Joshua~P Hecker and Melanie~E Moses.
\newblock Beyond pheromones: evolving error-tolerant, flexible, and scalable
  ant-inspired robot swarms.
\newblock \emph{Swarm Intelligence}, 9\penalty0 (1):\penalty0 43--70, 2015.

\bibitem[Wilensky(1997)]{wilensky1997netlogo}
Uri Wilensky.
\newblock Netlogo ants model.
\newblock \emph{Center for Connected Learning and Computer-Based Modeling,
  Northwestern University, Evanston, IL}, 1997.

\bibitem[Duan and Sun(2014)]{duan2014swarm}
Haibin Duan and Changhao Sun.
\newblock Swarm intelligence inspired shills and the evolution of cooperation.
\newblock \emph{Scientific reports}, 4\penalty0 (1):\penalty0 1--8, 2014.

\bibitem[Chevallier et~al.(2010)Chevallier, Paugam-Moisy, and
  Sebag]{chevallier2010spikeants}
Sylvain Chevallier, H{\'e}l{\`e}ne Paugam-Moisy, and Mich{\`e}le Sebag.
\newblock Spikeants, a spiking neuron network modelling the emergence of
  organization in a complex system.
\newblock In \emph{NIPS'2010}, pages 379--387, 2010.

\bibitem[Christensen and Dorigo(2006)]{christensen2006evolving}
Anders~Lyhne Christensen and Marco Dorigo.
\newblock Evolving an integrated phototaxis and hole-avoidance behavior for a
  swarm-bot.
\newblock In \emph{Artificial Life X: Proceedings of the Tenth International
  Conference on the Simulation and Synthesis of Living Systems. Cambridge: MIT
  Press. A Bradford Book}, pages 248--254, 2006.

\bibitem[Trianni and Nolfi(2009)]{trianni2009self}
Vito Trianni and Stefano Nolfi.
\newblock Self-organizing sync in a robotic swarm: a dynamical system view.
\newblock \emph{IEEE Transactions on Evolutionary Computation}, 13\penalty0
  (4):\penalty0 722--741, 2009.

\bibitem[Waibel et~al.(2009)Waibel, Keller, and Floreano]{waibel2009genetic}
Markus Waibel, Laurent Keller, and Dario Floreano.
\newblock Genetic team composition and level of selection in the evolution of
  cooperation.
\newblock \emph{IEEE transactions on Evolutionary Computation}, 13\penalty0
  (3):\penalty0 648--660, 2009.

\bibitem[Ericksen et~al.(2017)Ericksen, Moses, and
  Forrest]{ericksen2017automatically}
John Ericksen, Melanie Moses, and Stephanie Forrest.
\newblock Automatically evolving a general controller for robot swarms.
\newblock In \emph{2017 IEEE symposium series on computational intelligence
  (SSCI)}, pages 1--8. IEEE, 2017.

\bibitem[Francesca and Birattari(2016)]{francesca2016automatic}
Gianpiero Francesca and Mauro Birattari.
\newblock Automatic design of robot swarms: achievements and challenges.
\newblock \emph{Frontiers in Robotics and AI}, 3:\penalty0 29, 2016.

\bibitem[Trianni et~al.(2008)Trianni, Nolfi, and Dorigo]{trianni2008evolution}
Vito Trianni, Stefano Nolfi, and Marco Dorigo.
\newblock Evolution, self-organization and swarm robotics.
\newblock \emph{Swarm intelligence: introduction and applications}, pages
  163--191, 2008.

\bibitem[Yamazaki et~al.(2022)Yamazaki, Vo-Ho, Bulsara, and
  Le]{yamazaki2022spiking}
Kashu Yamazaki, Viet-Khoa Vo-Ho, Darshan Bulsara, and Ngan Le.
\newblock Spiking neural networks and their applications: A review.
\newblock \emph{Brain Sciences}, 12\penalty0 (7):\penalty0 863, 2022.

\bibitem[Nichols et~al.(2010)Nichols, McDaid, and Siddique]{nichols2010case}
Eric Nichols, LJ~McDaid, and NH~Siddique.
\newblock Case study on a self-organizing spiking neural network for robot
  navigation.
\newblock \emph{International Journal of Neural Systems}, 20\penalty0
  (06):\penalty0 501--508, 2010.

\bibitem[Beyeler et~al.(2015)Beyeler, Oros, Dutt, and Krichmar]{beyeler2015gpu}
Michael Beyeler, Nicolas Oros, Nikil Dutt, and Jeffrey~L Krichmar.
\newblock A gpu-accelerated cortical neural network model for visually guided
  robot navigation.
\newblock \emph{Neural Networks}, 72:\penalty0 75--87, 2015.

\bibitem[Nichols et~al.(2012)Nichols, McDaid, and
  Siddique]{nichols2012biologically}
Eric Nichols, Liam~J McDaid, and Nazmul Siddique.
\newblock Biologically inspired snn for robot control.
\newblock \emph{IEEE transactions on cybernetics}, 43\penalty0 (1):\penalty0
  115--128, 2012.

\bibitem[Basu et~al.(2022)Basu, Deng, Frenkel, and Zhang]{basu2022spiking}
Arindam Basu, Lei Deng, Charlotte Frenkel, and Xueyong Zhang.
\newblock Spiking neural network integrated circuits: A review of trends and
  future directions.
\newblock In \emph{2022 IEEE Custom Integrated Circuits Conference (CICC)},
  pages 1--8. IEEE, 2022.

\bibitem[Ottati et~al.(2023)Ottati, Gao, Chen, Brignone, Casu, Eshraghian, and
  Lavagno]{ottati2023spike}
Fabrizio Ottati, Chang Gao, Qinyu Chen, Giovanni Brignone, Mario~R Casu,
  Jason~K Eshraghian, and Luciano Lavagno.
\newblock To spike or not to spike: A digital hardware perspective on deep
  learning acceleration.
\newblock \emph{arXiv preprint arXiv:2306.15749}, 2023.

\bibitem[Putra and Shafique(2023)]{putra2023topspark}
Rachmad Vidya~Wicaksana Putra and Muhammad Shafique.
\newblock Topspark: a timestep optimization methodology for energy-efficient
  spiking neural networks on autonomous mobile agents.
\newblock \emph{arXiv preprint arXiv:2303.01826}, 2023.

\bibitem[Yegenoglu et~al.(2022)Yegenoglu, Subramoney, Hater, Jimenez-Romero,
  Klijn, Pérez~Martín, van~der Vlag, Herty, Morrison, and
  Diaz-Pier]{yegenoglu2022exploring}
Alper Yegenoglu, Anand Subramoney, Thorsten Hater, Cristian Jimenez-Romero,
  Wouter Klijn, Aarón Pérez~Martín, Michiel van~der Vlag, Michael Herty,
  Abigail Morrison, and Sandra Diaz-Pier.
\newblock Exploring parameter and hyper-parameter spaces of neuroscience models
  on high performance computers with learning to learn.
\newblock \emph{Frontiers in Computational Neuroscience}, 16, 2022.
\newblock ISSN 1662-5188.
\newblock \doi{10.3389/fncom.2022.885207}.
\newblock URL
  \url{https://www.frontiersin.org/articles/10.3389/fncom.2022.885207}.

\bibitem[Tisue and Wilensky(2004)]{tisue2004netlogo}
Seth Tisue and Uri Wilensky.
\newblock Netlogo: A simple environment for modeling complexity.
\newblock In \emph{International conference on complex systems}, volume~21,
  pages 16--21. Boston, MA, 2004.

\bibitem[Gewaltig and Diesmann(2007)]{gewaltig2007nest}
Marc-Oliver Gewaltig and Markus Diesmann.
\newblock Nest (neural simulation tool).
\newblock \emph{Scholarpedia}, 2\penalty0 (4):\penalty0 1430, 2007.

\bibitem[Fortin et~al.(2012)Fortin, {De Rainville}, Gardner, Parizeau, and
  Gagn\'e]{DEAP_JMLR2012}
F\'elix-Antoine Fortin, Fran\c{c}ois-Michel {De Rainville}, Marc-Andr\'e
  Gardner, Marc Parizeau, and Christian Gagn\'e.
\newblock {DEAP}: Evolutionary algorithms made easy.
\newblock \emph{Journal of Machine Learning Research}, 13:\penalty0 2171--2175,
  jul 2012.

\bibitem[Jordan et~al.(2018)Jordan, Ippen, Helias, Kitayama, Sato, Igarashi,
  Diesmann, and Kunkel]{jordan2018extremely}
Jakob Jordan, Tammo Ippen, Moritz Helias, Itaru Kitayama, Mitsuhisa Sato, Jun
  Igarashi, Markus Diesmann, and Susanne Kunkel.
\newblock Extremely scalable spiking neuronal network simulation code: from
  laptops to exascale computers.
\newblock \emph{Frontiers in neuroinformatics}, 12:\penalty0 2, 2018.

\bibitem[Gerstner et~al.(2014)Gerstner, Kistler, Naud, and
  Paninski]{gerstner2014neuronal}
Wulfram Gerstner, Werner~M Kistler, Richard Naud, and Liam Paninski.
\newblock \emph{Neuronal dynamics: From single neurons to networks and models
  of cognition}.
\newblock Cambridge University Press, 2014.

\bibitem[Denker et~al.(2018)Denker, Yegenoglu, and Grün]{elephant18}
M.~Denker, A.~Yegenoglu, and S.~Grün.
\newblock {C}ollaborative {HPC}-enabled workflows on the {HBP} {C}ollaboratory
  using the {E}lephant framework.
\newblock In \emph{Neuroinformatics 2018}, page P19, 2018.
\newblock \doi{10.12751/incf.ni2018.0019}.
\newblock URL
  \url{https://abstracts.g-node.org/conference/NI2018/abstracts#/uuid/023bec4e-0c35-4563-81ce-2c6fac282abd}.

\bibitem[Robinson et~al.(2005)Robinson, Jackson, Holcombe, and
  Ratnieks]{NegPheromone2005}
Elva Robinson, Duncan Jackson, M.~Holcombe, and Francis Ratnieks.
\newblock Insect communication: ‘no entry’ signal in ant foraging.
\newblock \emph{Nature}, 438:\penalty0 442, 12 2005.
\newblock \doi{10.1038/438442a}.

\bibitem[Ocker et~al.(2015)Ocker, Litwin-Kumar, and Doiron]{ocker2015self}
Gabriel~Koch Ocker, Ashok Litwin-Kumar, and Brent Doiron.
\newblock Self-organization of microcircuits in networks of spiking neurons
  with plastic synapses.
\newblock \emph{PLoS computational biology}, 11\penalty0 (8):\penalty0
  e1004458, 2015.

\bibitem[Kauffman et~al.(1993)]{kauffman1993origins}
Stuart~A Kauffman et~al.
\newblock \emph{The origins of order: Self-organization and selection in
  evolution}.
\newblock Oxford University Press, USA, 1993.

\bibitem[Staudinger et~al.(2018)Staudinger, Shutin, Man{\ss}, Viseras, and
  Zhang]{staudinger2018swarm}
Emanuel Staudinger, Dmitriy Shutin, Christoph Man{\ss}, Alberto Viseras, and
  Siwei Zhang.
\newblock Swarm technologies for future space exploration missions.
\newblock In \emph{ISAIRAS'18: FOURTEENTH INTERNATIONAL SYMPOSIUM ON ARTIFICIAL
  INTELLIGENCE, ROBOTICS AND AUTOMATION IN SPACE}, 2018.

\bibitem[Dorigo et~al.(2020)Dorigo, Theraulaz, and
  Trianni]{dorigo2020reflections}
Marco Dorigo, Guy Theraulaz, and Vito Trianni.
\newblock Reflections on the future of swarm robotics.
\newblock \emph{Science Robotics}, 5\penalty0 (49):\penalty0 eabe4385, 2020.

\end{thebibliography}
\end{document}